\documentclass[preprint,11pt,3p,authoryear]{elsarticle}
\usepackage{wrapfig}
\usepackage[T1]{fontenc}
\usepackage{graphicx}
\usepackage{subcaption}
\usepackage{amsmath,amssymb,amsfonts}
\usepackage{float}
\usepackage{hyperref}
\usepackage{color,xcolor}
\usepackage{booktabs,multirow}

\usepackage{mathtools}
\usepackage{natbib}
\usepackage{amssymb}
\usepackage{placeins}
\usepackage{needspace}
\usepackage{tabularx} 
\renewcommand{\arraystretch}{1.1} 
\usepackage{amsthm}
\usepackage{enumitem}
\newtheorem{definition}{Definition}
\usepackage{colortbl}
\usepackage{comment}
\usepackage{verbatim}
\usepackage{algorithm}
\usepackage{algpseudocode}
\usepackage{amsmath}  
\newcounter{ToDo}
\newcounter{gaocomm}
\newcounter{wangcomm}
\newcounter{Note}
\definecolor{blue-violet}{rgb}{0.00,0.75,0.90}
\definecolor{mygreen}{rgb}{0.0, 0.5, 0.0}
\definecolor{awesome}{rgb}{1.0, 0.13, 0.32}
\definecolor{bostonuniversityred}{rgb}{1.0, 0.0, 0.0}

\usepackage{graphicx} 

\usepackage{amssymb}
\usepackage{amsmath}
\usepackage{makecell}

\journal{Transportation Research Part C}

\begin{document}

\begin{frontmatter}



\title{Toward an Integrated Cross-Urban Accident Prevention System: A Multi-Task Spatial–Temporal Learning Framework for Urban Safety Management} 


\author[label1]{Jiayu Fang}
\ead{jfan4872@uni.sydney.edu.au}

\author[label1,label3]{Zhiqi Shao\corref{cor1}}
\ead{zhiqi.shao@sydney.edu.au}

\author[label2,label3]{Haoning Xi}
\ead{alice.xi@newcastle.edu.au}

\author[label1]{Boris Choy}
\ead{boris.choy@sydney.edu.au}

\author[label1]{Junbin Gao}
\ead{junbin.gao@sydney.edu.au}

\cortext[cor1]{Corresponding author.}

\affiliation[label1]{%
  organization={Discipline of Business Analytics, The University of Sydney Business School},%
  addressline={The University of Sydney},%
  city={Camperdown},%
  postcode={2006},%
  state={NSW},%
  country={Australia}%
}

\affiliation[label2]{%
  organization={Newcastle Business School},%
  addressline={The University of Newcastle},%
  city={Newcastle},%
  postcode={2300},%
  state={NSW},%
  country={Australia}%
}

\affiliation[label3]{%
  organization={Institute of Transport and Logistics Studies, The University of Sydney Business School},%
  addressline={The University of Sydney},%
  city={Camperdown},%
  postcode={2006},%
  state={NSW},%
  country={Australia}%
}



\begin{abstract}

The development of a cross-city accident prevention system is particularly challenging due to the heterogeneity, inconsistent reporting, and inherently clustered, sparse, cyclical, and noisy nature of urban accident data. These intrinsic data properties, combined with fragmented governance and incompatible reporting standards, have long hindered the creation of an integrated, cross-city accident prevention framework. To address this gap, we propose the Mamba Local-Attention Spatial–Temporal Network (\textsf{MLA-STNet}), a unified system that formulates accident risk prediction as a multi-task learning problem across multiple cities. \textsf{MLA-STNet} integrates two complementary modules: (i) the Spatio-Temporal Geographical Mamba-Attention (STG-MA), which suppresses unstable spatio-temporal fluctuations and strengthens long-range temporal dependencies; and (ii) the Spatio-Temporal Semantic Mamba-Attention (STS-MA), which mitigates cross-city heterogeneity through a shared-parameter design that jointly trains all cities while preserving individual semantic representation spaces.
We validate the proposed framework through 75 experiments under two forecasting scenarios, full-day and high-frequency accident periods, using real-world datasets from New York City and Chicago. Compared with the state-of-the-art baselines, \textsf{MLA-STNet} achieves up to 6\% lower RMSE, 8\% higher Recall, and 5\% higher MAP, while maintaining less than 1\% performance variation under 50\% input noise. These results demonstrate that \textsf{MLA-STNet} effectively unifies heterogeneous urban datasets within a scalable, robust, and interpretable Cross-City Accident Prevention System, paving the way for coordinated and data-driven urban safety management.
\end{abstract}




\end{frontmatter}
\section{Introduction}
Urban regions worldwide have invested heavily in accident prevention systems, deploying technologies such as smart traffic lights, roadside sensors, targeted patrols, and intersection redesigns~\citep{ESSA2020105713}. These efforts have yielded measurable safety gains within individual jurisdictions: for example, Hyderabad focuses on real-time crash detection to accelerate ambulance response~\citep{Shah2024}, while Los Angeles, Brisbane, and Taipei emphasize engineering interventions or sobriety checkpoints~\citep{Morrison2019Sobriety}. Yet, their effectiveness remains fragmented, as governance silos confine promising solutions within single jurisdictions rather than enabling a unified Cross-City Accident Prevention System. The core challenge is that such a system must operate in a multi-task prediction environment, since each city follows distinct reporting protocols, infrastructure patterns, and enforcement capacities, producing heterogeneous and often incompatible datasets. Beyond this cross-city heterogeneity, the intrinsic nature of accident data itself also poses fundamental obstacles, further increasing the difficulty of establishing an effective cross-city accident prevention system.

Unlike ordinary urban mobility prediction tasks where events are more evenly distributed, accident risks exhibit four intrinsic characteristics: they are spatially clustered, temporally sparse, cyclical, and noisy. First, crashes concentrate heavily at intersections, with about 25\% of fatal and nearly 50\% of injury crashes occurring there, and over 40\% of all reported crashes happening in or near such locations \citep{FHWA2020-intersection-safety, FHWA2020-intersection-crash-rate}. Second, crashes are extremely sparse compared to the vast number of vehicle trips, which makes stable pattern learning difficult. Third, risks follow clear temporal cycles, spiking during commuting peaks and rising disproportionately at night due to visibility loss and impaired driving~\citep{NHTSA2007-nighttime, INRIX2018-scorecard}. Finally, accident records are often noisy, with missing or inconsistent reports obscuring true risk distributions. Together, these characteristics substantially increase the difficulty of developing a reliable cross-city accident prevention system. In summary, accident risk prediction faces intertwined challenges of (1) fragmented governance across cities, (2) heterogeneous and effectively multi-task datasets shaped by differing reporting protocols and infrastructures, (3) clustered, sparse, cyclical, and incomplete data distributions, and (4) systemic propagation of prediction errors into social and economic costs. These challenges motivate our data analysis in Section~\ref{cross}, where we empirically demonstrate the clustered, cyclical, and incomplete nature of accident risk data.

To directly tackle the above limitations, we propose Mamba Local-Attention Spatial–Temporal Net-work (\textsf{MLA-STNet}), a unified Cross-City Accident Prevention System that formulates accident forecasting as a multi-task prediction problem. By consolidating heterogeneous accident records into a single framework, \textsf{MLA-STNet} jointly learns from multiple related city-specific tasks, capturing both cross-city regularities and localized variations. It integrates spatio-temporal attention with multiple semantic graph structures to address spatial clustering, sparsity, temporal periodicity, and reporting noise, thereby overcoming data fragmentation and governance silos. This unified design benefits multiple stakeholders: agencies can coordinate resources across jurisdictions, planners can more reliably identify hazardous locations, emergency services gain stable hotspot predictions, policy makers and insurers obtain stronger evidence for risk management, and the public ultimately benefits from safer urban mobility. The major contributions of this paper are summarized as follows:

\begin{itemize}
    \item To the best of our knowledge, we propose the first Cross-City Accident Prevention System that is served by the model \textsf{MLA-STNet}, which jointly learns heterogeneous patterns across cities while preserving local variations, enabling coordinated resource allocation, hazard identification, faster emergency response, and safer mobility.

    \item We design the Spatio-Temporal Geographical Mamba-Attention (STG-MA) module to handle clustered, sparse, cyclical, and noisy accident data, filtering unreliable signals and strengthening long-range temporal dependencies.

    \item We introduce the Spatio–Temporal Semantic Mamba-Attention (STS-MA) module to address cross-city heterogeneity through a \textit{parameter-sharing framework}, where global parameters are shared across cities while each city independently learns its own semantic matrices, enabling collaborative optimization and preserving city-specific representations.

    \item We validate \textsf{MLA-STNet} through 75 experiments across two tasks: all-day forecasting and forecasting during high-frequency accident periods. Our experiments involve five state-of-the-art (SOTA) baselines on two real-world datasets, Chicago and New York City (NYC): (1) compared with the SOTA baseline, the \textsf{MLA-STNet} (multi-task) reduces RMSE by up to 6\%, improves Recall by up to 8\%, and increases MAP by over 5\%, while even the single-task variant consistently surpasses all baselines; and (2) \textsf{MLA-STNet} remains robust under noisy conditions, with RMSE, Recall, and MAP changing less than 1\% when noise intensity increases up to 0.5\%.

\end{itemize}

The remainder of this paper is organized as follows. Section~\ref{s1} reviews the related literature. Section~\ref{data} presents the dataset description and exploratory analytics. Section~\ref{s3} introduces the notations and formulates the problem statement. Section~\ref{s4} details the architecture of the proposed \textsf{MLA-STNet} model. Section~\ref{s5} reports the experimental results and performance evaluation. Section~\ref{s6} discusses implementation details and provides further insights. Finally, Section~\ref{s7} concludes the paper and outlines potential directions for future research.

\section{Related Works}
\label{s1}

Despite decades of efforts to improve road safety, accident prevention remains hindered by the fragmented nature of current predictive systems. Crashes emerge from a web of interacting factors that unfold across space, time, and data quality, yet most existing models approach this challenge in a narrowly defined, single-task manner. For example, Bayesian Dynamic Logistic Regression (BDLR)~\citep{yang2018bayesian} has been used to update risk estimates under incomplete data, but it struggles to incorporate spatial or temporal dynamics. Point-of-interest (POI)--based clustering methods~\citep{jia2018traffic} identify recurrent hotspots, yet they often ignore temporal evolution and the role of contextual features. The Imbalance-Aware Fusion Framework (LAFF)~\citep{abou2020real} improves recall of rare crash events, but it still cannot generalize across diverse environments. Similarly, Deep Convolutional Generative Adversarial Network (DCGAN)~\citep{cai2020real} captures cyclical variations, but they remain limited when data sparsity and cross-city transfer are considered. Other deep learning models, including Transformer-based architectures such as CCDSReFormer \citep{SHAO2025100189} and STAEFormer \citep{staeformer}, have improved spatial–temporal forecasting through attention mechanisms. However, these models generally require separate attention modules for spatial and temporal contexts, leading to increased computational cost and reduced scalability in real-time urban applications.
More recently, several studies have begun exploring state-space–based approaches such as Mamba \citep{SHAO2024102872, shao2024stmamba}, which can capture dynamic spatial dependencies and temporal delays more efficiently.
Yet, these Mamba extensions remain confined to single-city or mode-specific tasks (e.g., traffic flow or passenger demand), leaving cross-city accident prediction, where heterogeneity and data irregularity are most pronounced, largely unexplored. While each of these methods has advanced prediction within its specific dimension, their reliance on isolated city datasets makes them inherently fragmented and unable to integrate multiple risks simultaneously. In practice, however, accident prevention is never a single-task problem: cities must coordinate urban mobility management, emergency response, and medical resources across jurisdictions. Fragmented, city-specific models fail to support such collaboration, slowing down joint decision-making, duplicating interventions, and undermining public trust.

\begin{figure}[htbp!]
    \centering
    \includegraphics[width=\textwidth]{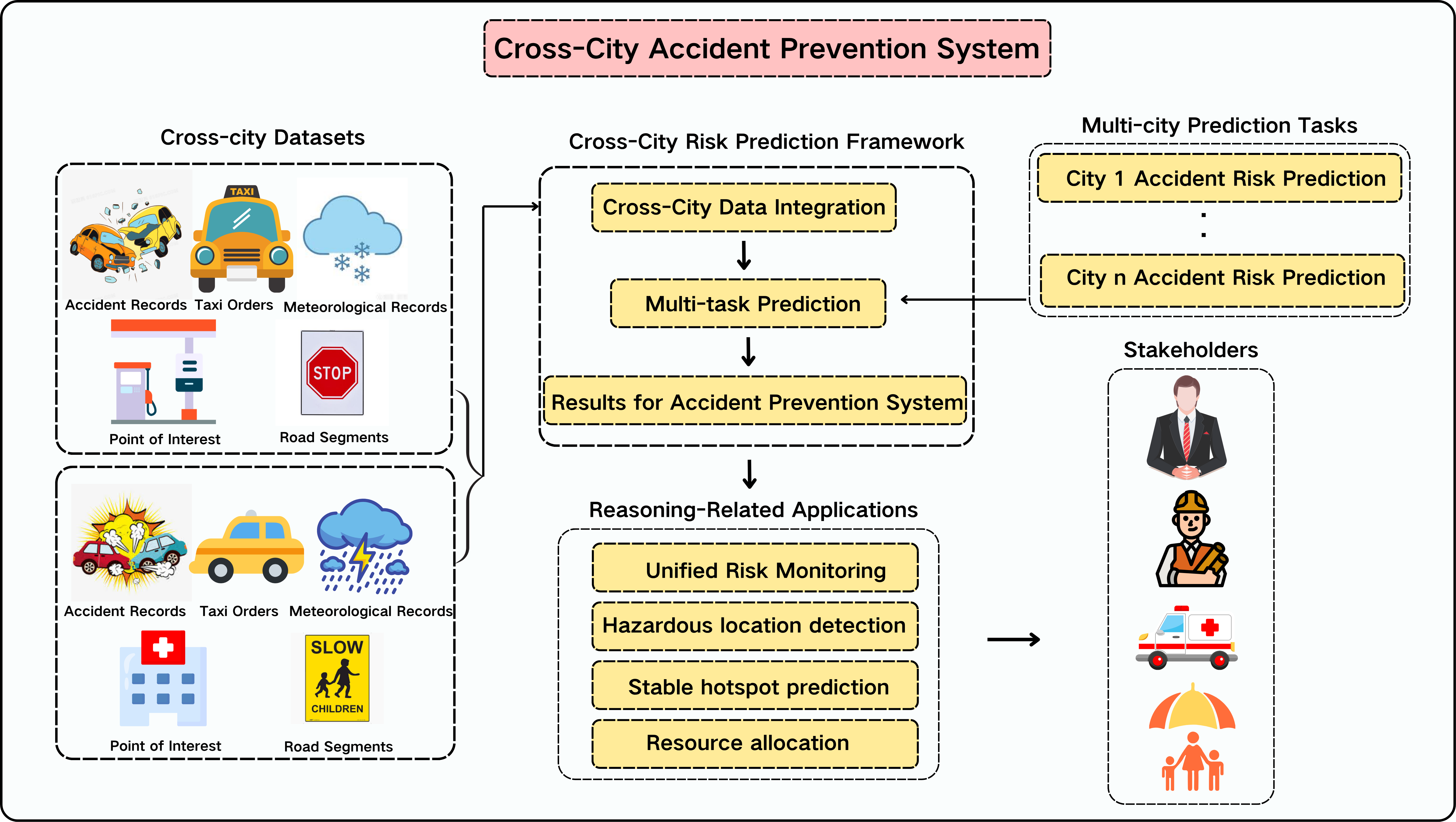} 
    \caption{Overall pipeline of the Cross-City Accident Prevention System, showing the flow from cross-city datasets to the risk prediction framework and finally to prediction tasks and reasoning-related applications.}

    \label{aps}
\end{figure}

Recognizing the limitations of single-task models, researchers have increasingly turned to multi-task learning to capture the interdependencies of urban mobility and safety. Initial progress was made in passenger demand and travel behavior, where joint frameworks were developed to predict ride-hailing demand across multiple modes, exemplified by the Deep Multi-Task Multi-Graph Learning Approach (DMT-MGL)~\citep{ke2021joint}. Building on this, the Personalized Travel Behavior Transformer with Multi-gate Mixture-of-Experts (PTBformer-MMoE)~\citep{xi2025multi}, a Transformer-based architecture with mixture-of-experts, was proposed to improve personalized periodic travel forecasts, and more recently, Spatial–Temporal Large Language Model with Denoising Diffusion (STLLM-DF)\citep{shao2025spatial} has been applied to enhance multimodal transport prediction~. In parallel, STDAtt-Mamba \citep{SHAO2025104282} proposed a spatial–temporal dynamic attention–based Mamba model for multi-type passenger demand prediction, showcasing the potential of Mamba state-space modeling for capturing dynamic dependencies across time and space. Encouraged by these advances, accident prediction also began to adopt multi-task paradigms: Multi-task learning Graph-based Network (MT-GN)~\citep{tran2023area} was introduced to classify jointly and regress incident duration, deep neural networks were extended to simultaneously predict multiple dimensions of accident severity through a Multi-task Deep Neural Network (MT-DNN) framework~\citep{yang2022predicting}, and the Uncertainty-Aware Traffic Accident Risk Prediction (TarU)~\citep{zhang2025uncertainty} was developed to improve robustness in accident risk prediction, effectively addressing dual imbalances and sparse supervision through hypergraph-enhanced contrastive learning. Nevertheless, despite these advances, accident forecasting still faces two fundamental challenges. First, accident data itself exhibits four intrinsic properties: it is spatially clustered, temporally periodic, extremely sparse, and frequently noisy or incomplete. These characteristics make stable pattern extraction difficult and increase the risk of biased or unreliable forecasts, often resulting in false alarms or misleading risk assessments. Second, urban environments are highly heterogeneous, differing in infrastructure, reporting standards, enforcement intensity, and mobility behaviors, leading to inconsistent emergency responses, severe misallocation of resources, and ultimately a diminished capacity to prevent accidents effectively. The comparative summary of representative models with respect to these intrinsic data properties, multi-task capabilities, and cross-city prediction is presented in Table~\ref{tab:related_models}. These challenges highlight the urgent need for a unified framework that can integrate multi-city accident data and support cross-city risk prediction (See Fig.~\ref{aps}).

Motivated by these challenges, we propose \textsf{MLA-STNet}, a model that can be easily adopted in a unified Cross-City Accident Prevention System formulated within a multi-task prediction environment. \textsf{MLA-STNet} contains two core modules. The Spatio-Temporal Geographical Mamba-Attention (STG-MA) module addresses the intrinsic characteristics of accident data, spatial clustering, temporal periodicity, sparsity, and noise, by combining Mamba-based long-range temporal modeling with localized masked attention to stabilize predictions and suppress unreliable signals. The Spatio-Temporal Semantic Mamba-Attention (STS-MA) module, in turn, tackles cross-city heterogeneity by constructing heterogeneous semantic graphs that integrate roadway topology, mobility demand, meteorological factors, and accident records. While each city maintains its own semantic graph to preserve local characteristics, Mamba kernels provide a shared parameterization that captures common accident-inducing patterns across cities. This dual mechanism allows \textsf{MLA-STNet} to respect city-specific contexts and exploit cross-city regularities simultaneously. Together, these two modules enable \textsf{MLA-STNet} to effectively address data irregularities and urban heterogeneity, establishing the first scalable and reliable Cross-City Accident Prevention System.

\begin{table}[htbp]
\centering
\caption{Comparison of representative models with respect to intrinsic accident data properties, multi-task capability, and cross-city prediction.}
\label{tab:related_models}
\footnotesize
\begin{tabular}{lcccccc}
\toprule
\textbf{Model name} &
\makecell{\textbf{Spatial}\\\textbf{clustering}} &
\makecell{\textbf{Temporal}\\\textbf{periodicity}} &
\makecell{\textbf{Sparsity}\\\textbf{handling}} &
\makecell{\textbf{Noise}\\\textbf{robustness}} &
\makecell{\textbf{Multi}\\\textbf{task}} &
\makecell{\textbf{Cross}\\\textbf{city}} \\
\midrule
BDLR~\citep{yang2018bayesian} & $\times$ & $\checkmark$ & $\times$ & $\times$ & $\times$ & $\times$ \\
\midrule
POI-based~\citep{jia2018traffic} & $\checkmark$ & $\times$ & $\times$ & $\times$ & $\times$ & $\times$ \\
\midrule
IAFF~\citep{abou2020real} & $\times$ & $\times$ & $\checkmark$ & $\times$ & $\times$ & $\times$ \\
\midrule
DCGAN~\citep{cai2020real} & $\times$ & $\checkmark$ & $\times$ & $\times$ & $\times$ & $\times$ \\
\midrule
DMT-MGL~\citep{ke2021joint} & $\checkmark$ & $\checkmark$ & $\checkmark$ & $\times$ & $\checkmark$ & $\times$ \\
\midrule
PTBformer-MMoE~\citep{xi2025multi} & $\times$ & $\checkmark$ & $\times$ & $\times$ & $\checkmark$ & $\times$ \\
\midrule
STLLM-DF~\citep{shao2025spatial} & $\checkmark$ & $\checkmark$ & $\checkmark$ & $\checkmark$ & $\checkmark$ & $\times$ \\
\midrule
MT-GN~\citep{tran2023area} & $\checkmark$ & $\checkmark$ & $\times$ & $\times$ & $\checkmark$ & $\times$ \\
\midrule
MT-DNN~\citep{yang2022predicting} & $\times$ & $\checkmark$ & $\times$ & $\times$ & $\checkmark$ & $\times$ \\
\midrule
TarU~\citep{zhang2025uncertainty} & $\checkmark$ & $\times$ & $\checkmark$ & $\checkmark$ & $\checkmark$ & $\times$ \\
\midrule
\rowcolor{gray!25} \textbf{\textsf{MLA-STNet} (Ours)} & $\checkmark$ & $\checkmark$ & $\checkmark$ & $\checkmark$ & $\checkmark$ & $\checkmark$ \\
\bottomrule
\end{tabular}
\end{table}

\section{Dataset Description and Analytics}
\label{data}
\subsection{Dataset Description}

Table \ref{tab:dataset_stats} summarizes the key characteristics of the New York City (NYC) and Chicago datasets used in this study. The NYC data were obtained from the City of New York’s Open Data portal\footnote{\url{https://data.cityofnewyork.us/Public-Safety/Motor-Vehicle-Collisions-Crashes/h9gi-nx95/about_data}}
, while the Chicago data were sourced from the U.S. Accidents dataset on Kaggle\footnote{\url{https://www.kaggle.com/datasets/sobhanmoosavi/us-accidents}}
. Both datasets integrate heterogeneous urban information, including traffic accident records, taxi trip data, road network attributes, and meteorological observations, thereby providing comprehensive representations of urban mobility and environmental context essential for cross-city accident risk modeling.

The New York City (NYC) dataset~\citep{Wang2021GSNetLS} comprises five data modalities: (1) \textit{traffic accident records}, which form the prediction target and contain approximately 147,000 crashes with detailed spatial–temporal annotations; (2) \textit{point-of-interest (POI)} data covering 15,625 regions, capturing residential, commercial, and cultural land-use structures; (3) \textit{taxi trip records} consisting of about 173 million orders that reflect fine-grained population mobility patterns; (4) \textit{road segment information} describing 103,000 transportation links; and (5) \textit{meteorological data} with 8,760 hourly weather entries, including precipitation and temperature, providing important environmental context. Despite its richness, accident events in this dataset are highly sparse compared with mobility volumes, posing substantial challenges for reliable pattern extraction and prediction.

The Chicago dataset~\citep{Wang2021GSNetLS} follows a similar structure but includes only four modalities, as POI data are unavailable. It contains 44,000 accident records from February to September 2016, 1.74 million taxi trip records, 56,000 road segments, and 5,832 meteorological entries. Compared with the NYC dataset, it is smaller in temporal span, lower in scale, and semantically incomplete due to missing POI information. These discrepancies reflect common challenges in multi-city traffic data analysis—specifically, variations in temporal coverage, data sparsity, and feature availability. Such heterogeneity underscores the need for a unified Cross-City Accident Prevention System capable of integrating diverse urban datasets while maintaining consistent predictive performance across cities.

\begin{table}[ht]
\centering
\footnotesize
\caption{Statistics of the New York City (NYC) and Chicago datasets.}

\label{tab:dataset_stats}
\begin{tabular}{l|c|c}
\hline
\textbf{Dataset} & \textbf{New York City (NYC)} & \textbf{Chicago} \\
\hline
Time span & 2013/01/01--2013/12/31 & 2016/02/01--2016/09/30 \\
Time granularity & 1h & 1h \\
Traffic accidents & 147k & 44k \\
Taxi orders & 173{,}179k & 1{,}744k \\
POI & 15{,}625 & None \\
Road segments & 103k & 56k \\
Meteorological records & 8{,}760 & 5{,}832 \\
\hline
\end{tabular}
\end{table}

\subsection{Cross-City Exploratory Analysis and Data Characteristics}
\label{cross}

Strong cyclic temporal regularities, clustered with sparse distributions, cross-city heterogeneity, and substantial data noise hinder accident risk prediction across metropolitan areas.  
To empirically examine these challenges, we analyze multi-year accident records from New York City (NYC) and Chicago. The analysis reveals how spatial clustering, temporal periodicity, and data incompleteness jointly shape the difficulty of developing a unified and reliable cross-city accident prevention system.

\begin{figure}[htbp!]
    \centering
    \includegraphics[width=\textwidth]{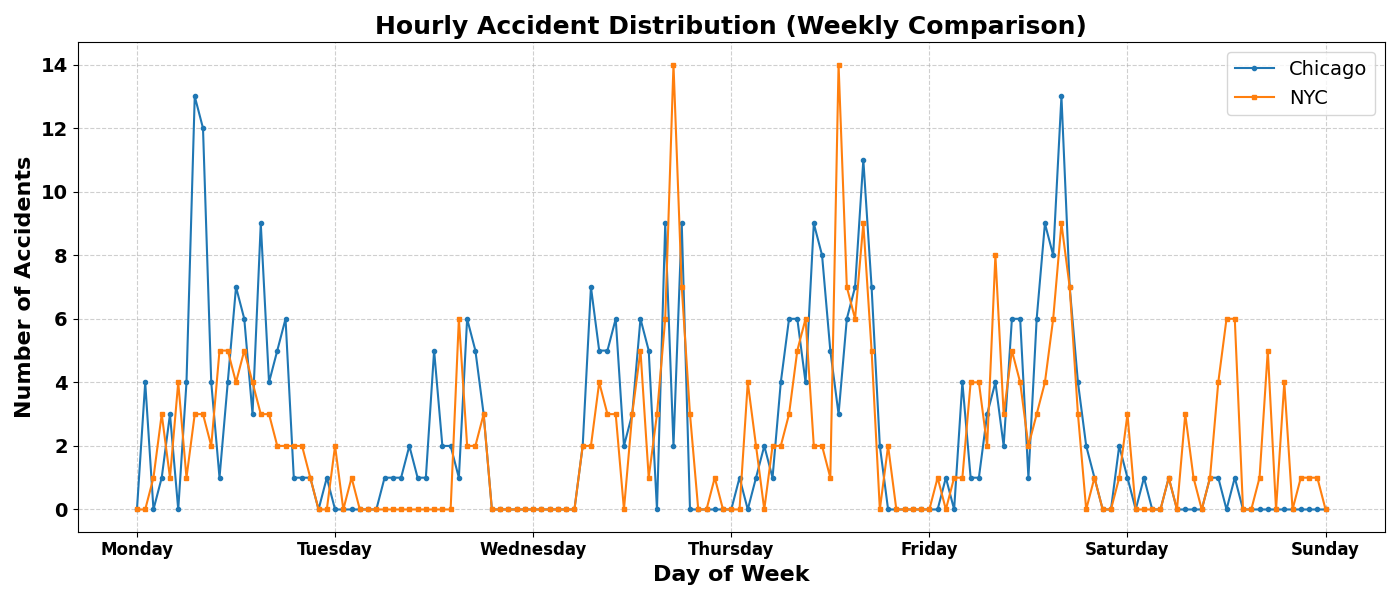}
    \caption{Hourly accident distribution over a week in Chicago and New York City (NYC).}
    \label{oneweek}
\end{figure}

\paragraph{Cyclic temporal regularities analysis}
Fig.~\ref{oneweek} illustrates the recurrent weekly rhythms of accident occurrences in New York City (NYC) and Chicago, revealing distinct temporal profiles that reflect city-specific mobility dynamics and commuting behaviors. 
NYC displays two pronounced daily peaks around 8:00–9:00 and 16:00–18:00 that persist into weekends, reflecting continuous commuting and leisure mobility. Chicago, however, shows a sharper evening peak and weaker morning activity, with flattened weekend curves. These findings verify the cyclical yet city-specific nature of temporal risks, confirming the need for flexible temporal modeling.

\begin{figure}[htbp!]
    \centering
    \includegraphics[width=1\textwidth]{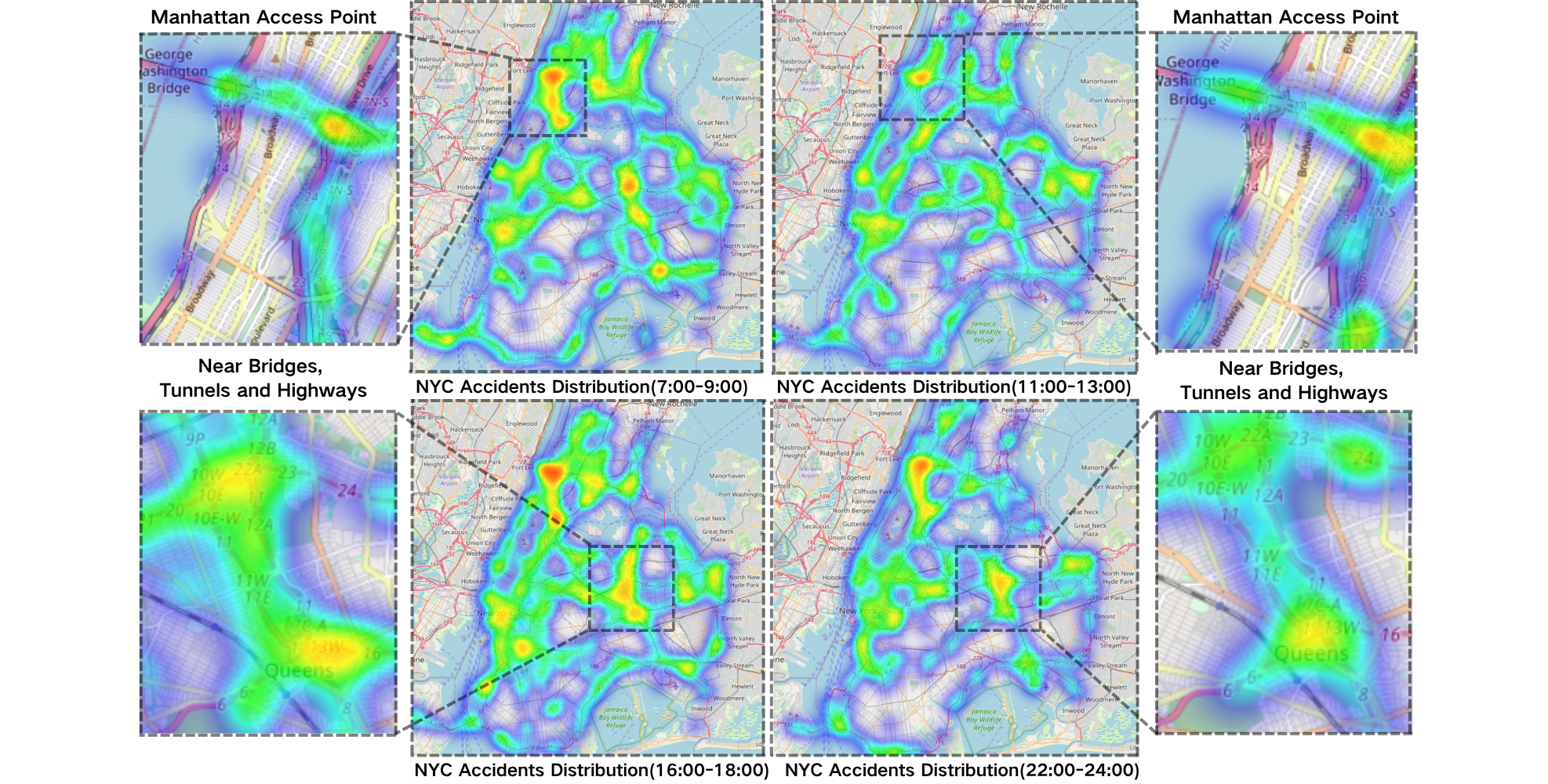}
    \caption{Spatial distribution of accidents across four time periods in New York City (NYC).}
    \label{fig:NYC_risk}
\end{figure}

\begin{figure}[htbp!]
    \centering
    \includegraphics[width=0.85\textwidth]{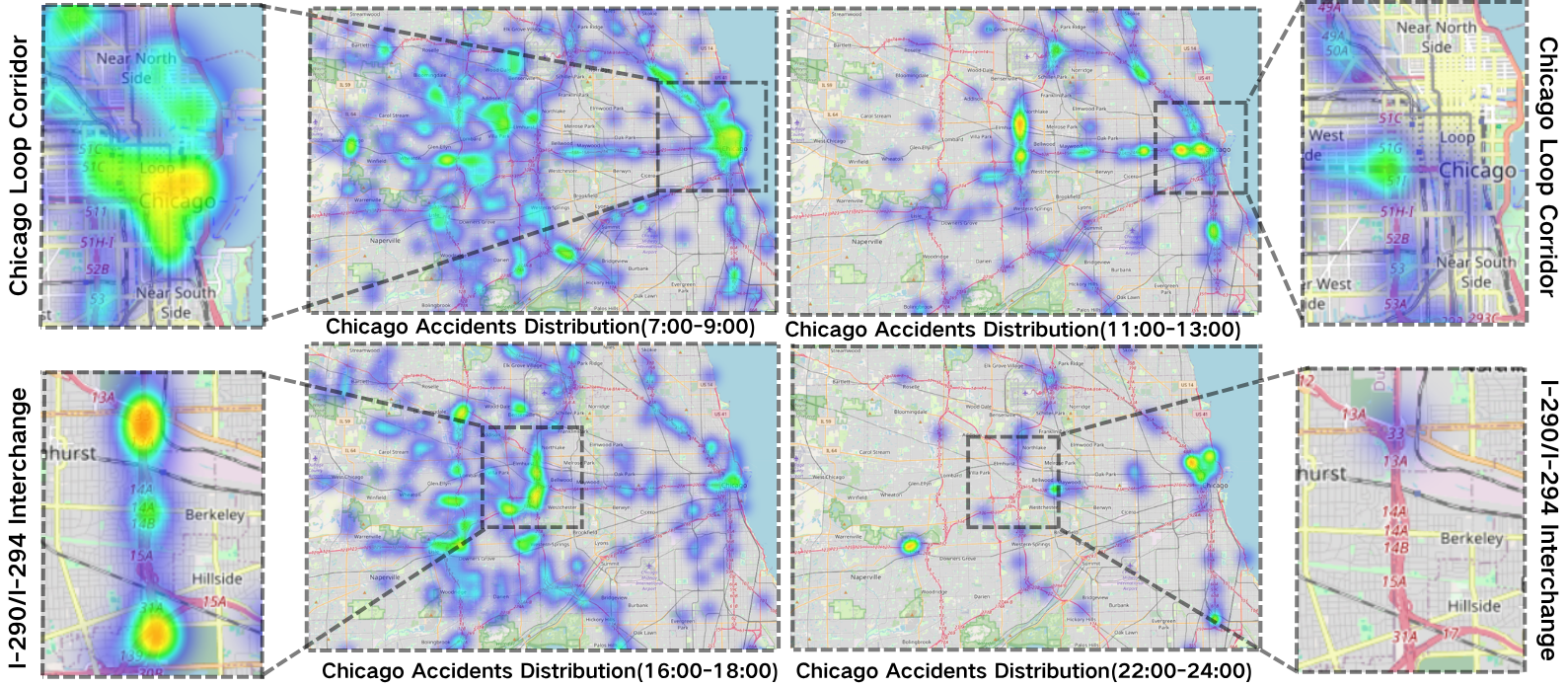}
    \caption{Spatial distribution of accidents across four time periods in Chicago.}
    \label{fig:CHI_risk}
\end{figure}

\paragraph{Clustered with sparse distributions analysis}
Fig.~\ref{fig:NYC_risk}–\ref{fig:CHI_risk} present the spatial distribution of accidents in New York City (NYC) and Chicago, highlighting the presence of highly clustered yet spatially sparse risk patterns across both urban networks.
In NYC, hotspots concentrate near Manhattan access points, including bridges, tunnels, and expressway connectors, forming a bridge–tunnel radial structure. In Chicago, risks cluster around the downtown Loop and interchanges such as I-290/I-294, following a hub-and-spoke highway design.  
These contrasting spatial logics highlight heterogeneous risk landscapes that complicate generalization across cities.

\begin{figure}[htbp!]
    \centering
    \begin{subfigure}[b]{0.49\textwidth}
        \centering
        \includegraphics[width=\textwidth]{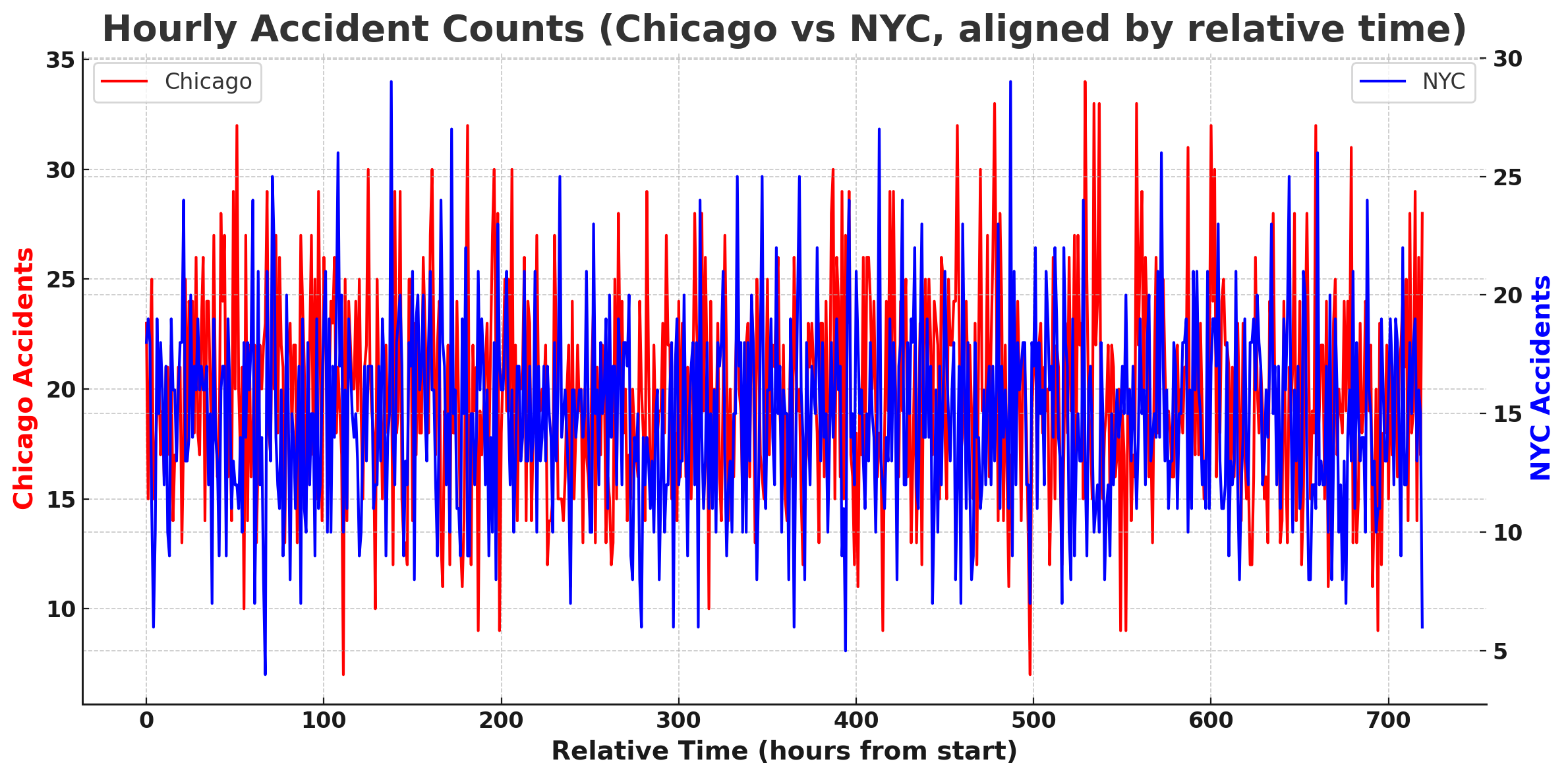}
        \caption{Hourly accident trends in Chicago and NYC.}
    \end{subfigure}
    \hfill
    \begin{subfigure}[b]{0.49\textwidth}
        \centering
        \includegraphics[width=\textwidth]{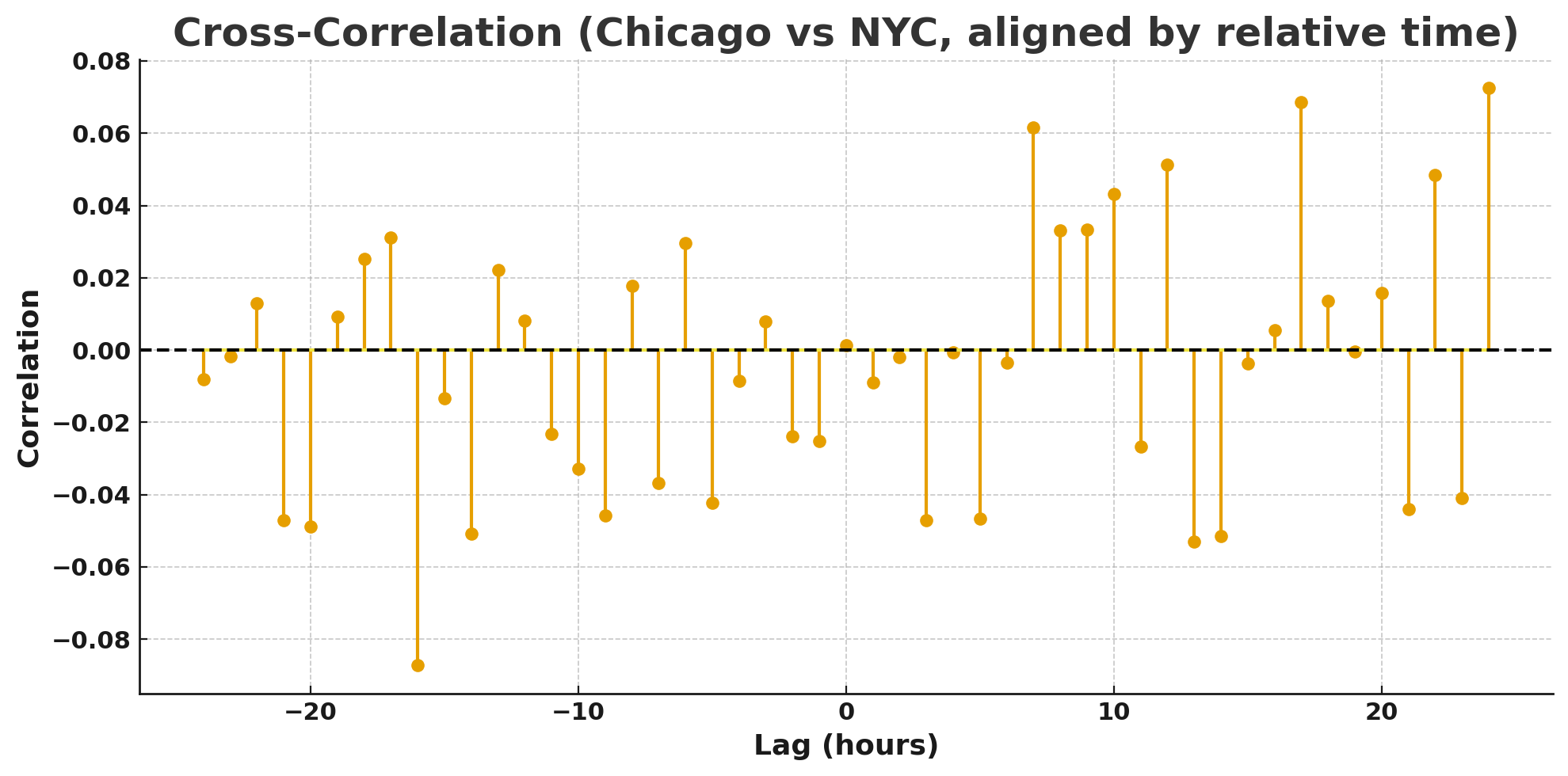}
        \caption{Cross-correlation of accident series.}
    \end{subfigure}
    \caption{Cross-city relationship analysis between Chicago and NYC.}
    \label{fig:Correlation}
\end{figure}

\paragraph{Cross-city heterogeneity analysis}
Fig.~\ref{fig:Correlation} illustrates the cross-city alignment of hourly accident series, conducted to evaluate the temporal heterogeneity between Chicago and New York City (NYC) by mapping both datasets onto a unified relative timeline. 
Chicago exhibits higher baseline frequencies with smoother variation, whereas NYC shows lower counts but stronger oscillations.  
Cross-correlation over $\pm24$-hour lags reveals weak positive peaks when NYC lags Chicago, hinting at shared exogenous drivers such as regional weather systems, while rolling correlations oscillate between positive and negative regimes.  
These patterns confirm that cross-city dependencies are dynamic and non-stationary, reinforcing the need for adaptive, multi-task learning.

\begin{figure}[!htbp]
    \centering
    \begin{subfigure}[b]{0.6\textwidth}
        \centering
        \includegraphics[width=\textwidth]{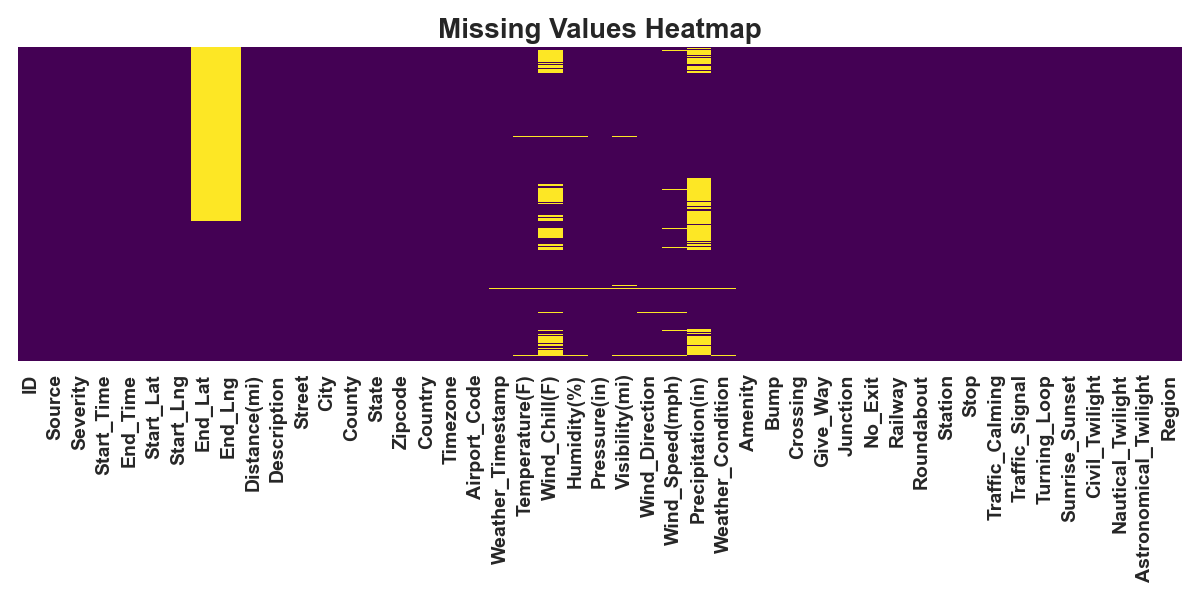}
        \caption{Missing-value heatmap.}
    \end{subfigure}
    \vskip\baselineskip
    \begin{subfigure}[b]{0.6\textwidth}
        \centering
        \includegraphics[width=\textwidth]{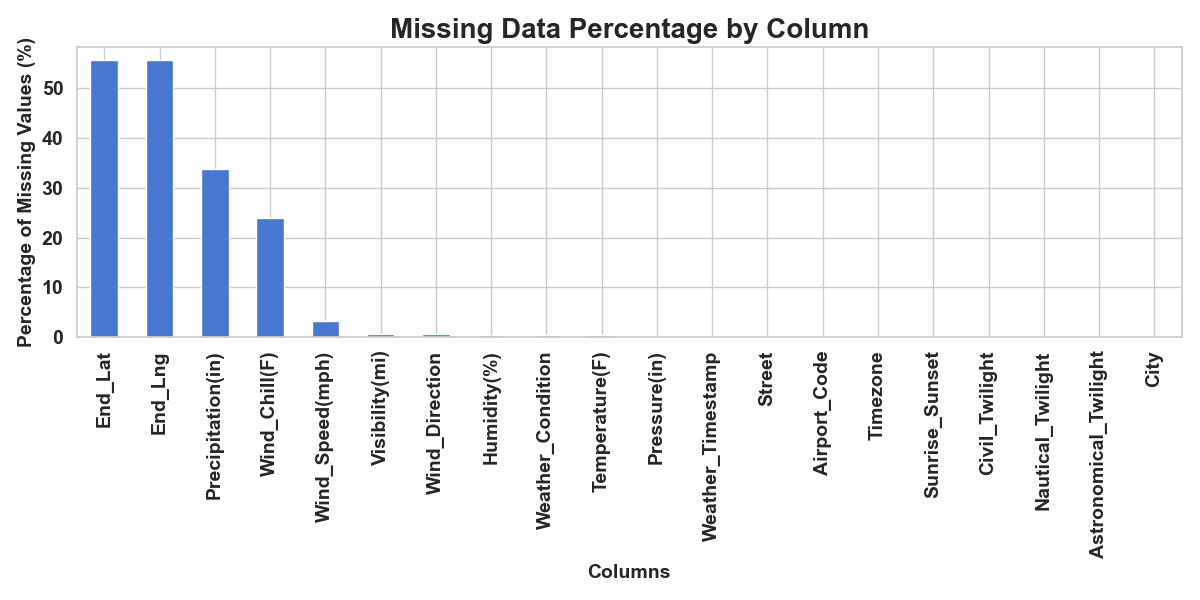}
        \caption{Proportion of missing values.}
    \end{subfigure}
    \vskip\baselineskip
    \begin{subfigure}[b]{0.6\textwidth}
        \centering
        \includegraphics[width=\textwidth]{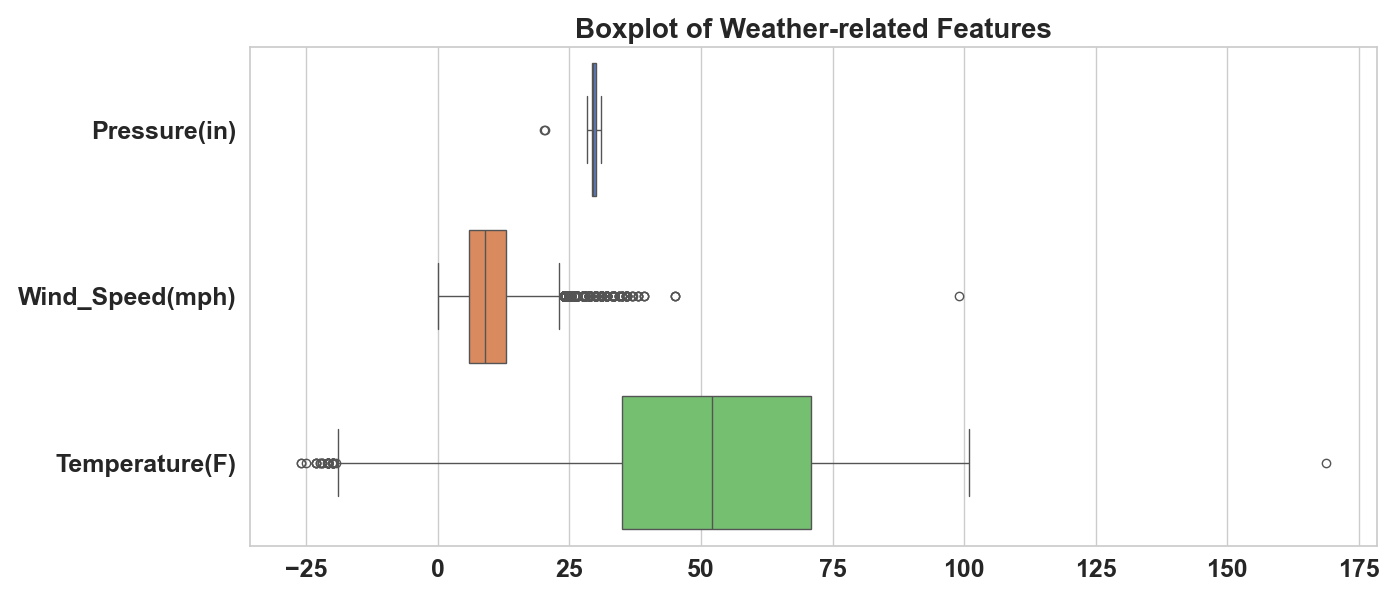}
        \caption{Outliers in weather variables.}
    \end{subfigure}
    \caption{Visualization of data incompleteness and noise in the Chicago dataset.}
    \label{fig:data_quality}
\end{figure}

\paragraph{Data incompleteness and noise analysis}
Fig.~\ref{fig:data_quality} presents the data quality assessment of the accident records, revealing considerable missing values and sensor noise that distort the observed spatial–temporal patterns of accident risk. In the Chicago dataset, large gaps appear in end-point coordinates and precipitation variables, and meteorological attributes (temperature, wind speed, pressure) show extreme outliers indicative of sensor or reporting errors. Such deficiencies bias learning and inflate uncertainty if left untreated. To mitigate these issues, we adopt a robustness-oriented preprocessing strategy: (i) a precomputed \emph{risk mask} excludes missing or unreliable entries from the loss and gradient updates during both training and inference; (ii) \emph{multi-view structural priors}, including risk co-occurrence similarity, road-network adjacency, and POI similarity, supply compensatory spatial/semantic signals; and (iii) a \emph{learnable adaptive similarity} matrix, refined end-to-end, denoises and reconstructs stable spatio-temporal relations. Together, these components reduce the influence of corrupted observations and support reliable cross-city learning.

In summary, the analysis reveals clustered and cyclical accident patterns, pronounced spatial–temporal heterogeneity, and significant data noise. These characteristics motivate the design of \textsf{MLA-STNet}, a unified cross-city forecasting model that learns common structures while remaining robust to missing and unreliable information.

\subsection{Data Pre-processing}
To account for the temporal discrepancy between the New York City (NYC) dataset (2013) and the Chicago dataset (2016), we apply a relative-time alignment procedure that standardizes the temporal reference frames of both datasets for consistent cross-city analysis. Specifically, we re-index all accident records in each dataset using an hourly index that starts from the first Monday 00:00 of their respective observation periods. In this way, both datasets follow a common weekly cycle (Monday to Sunday), allowing events to be compared across corresponding time slots in a typical week. After establishing this weekly synchronization, we further ensure that all features are aligned at an hourly granularity, so that accident events and explanatory variables (e.g., weather, traffic flow, or temporal indicators) are temporally consistent across both cities. As a result, the aligned datasets share a unified relative time axis, where each hour within the week represents the same temporal position, facilitating coherent spatio-temporal modeling and cross-city analysis.


A systematic preprocessing pipeline was then developed to transform the time-aligned datasets into a format suitable for one-step-ahead risk forecasting. The temporally synchronized accident records, originally stored in serialized \texttt{.pkl} files, were parsed and standardized into structured arrays that include accident risk values, temporal indicators, meteorological attributes, and mobility-related variables. Together, these components provide the spatio–temporal and contextual features required by the proposed \textsf{MLA-STNet} model.

Spatial structures were represented through multiple complementary supports. A \textit{risk mask} was generated to identify valid observations and exclude missing or unreliable entries during optimization. Three types of adjacency matrices were then constructed: (i) a \textit{road-network adjacency} capturing physical connectivity between nodes, (ii) a \textit{risk similarity} matrix encoding historical co-occurrence correlations among regions, and (iii) a \textit{POI similarity} matrix reflecting functional proximity derived from point-of-interest distributions. A mapping file was further produced to align grid-based features with corresponding graph nodes, thereby maintaining geometric consistency between spatial and semantic domains.

To formulate the supervised learning samples, a sliding-window strategy was applied, where the previous 12 time steps served as model inputs and the subsequent 1 time step as the prediction target. The processed samples were partitioned into training, validation, and testing sets with a ratio of 70\%, 10\%, and 20\%, respectively. To guarantee reproducibility and accelerate future data loading, all intermediate outputs were cached in NumPy (\texttt{.npy}) format. 

Overall, this preprocessing framework transforms heterogeneous accident records into unified spatio–temporal representations, embeds them within consistent graph structures, and ensures computational efficiency for subsequent training and evaluation of the \textsf{MLA-STNet} model.


\section{Preliminary}
\label{s3}

\subsection{Notations and Definitions}


We consider a multi-city accident forecasting problem involving $C$ heterogeneous urban environments.  
Each city $c \in \{1, \ldots, C\}$ possesses a unique spatial partition and road network structure.  
At every aligned time index $t \in \{1, \ldots, T\}$, we observe both grid-level and node-level features.  
A dataset sample $k \in \{1, \ldots, K\}$ aggregates multi-city observations over a sliding temporal window of length $T$,  
so that each sample contains synchronized spatio–temporal sequences for all $C$ cities.\footnote{When timestamps differ across cities, alignment is performed by nearest valid time within a tolerance of $\pm \Delta$.}

\paragraph{Geographical input} 
For each city $c$, the grid-based observation sequence in sample $k$ is denoted as
\[
X^{geo,(k,c)} = \{\mathbf{x}_t^{(m,k,c)}\}_{t=1}^{T}, \quad m \in [1,M_c], 
\quad X^{geo,(k,c)} \in \mathbb{R}^{T \times M_c \times F_{geo}},
\]
where $T$ is the input timestep length, and $F_{\text{geo}}$ is the feature dimension of each grid cell.
For city $c$, the spatial domain is discretized into a grid of size $W_c \times H_c$, 
where $W_c$ and $H_c$ denote the north--south and west--east resolutions, respectively.
Each feature vector $\mathbf{x}_t^{(m,k,c)} \in \mathbb{R}^{F_{\mathrm{geo}}}$  
represents the attributes of cell $m$ at time $t$, such as traffic volume, weather, and historical risk indicators.

\paragraph{Semantic input} 
Similarly, for each city $c$, the node-based observation sequence in sample $k$ is
\[
X^{sem,(k,c)} = \{\mathbf{x}_t^{(n,k,c)}\}_{t=1}^{T}, \quad n \in [1,N_c], 
\quad X^{sem,(k,c)} \in \mathbb{R}^{T \times N_c \times F_{sem}},
\]
where $N_c$ is the number of nodes in the road network of city $c$, 
and $F_{sem}$ is the feature dimension at each node. 

The heterogeneous spatial relations among nodes are encoded by a set of city-specific adjacency matrices:

\[
\mathcal{A}^{(c)} = \{A^{road,(c)}, A^{risk,(c)}, A^{poi,(c)}\}, \quad A^{(\cdot),(c)} \in \mathbb{R}^{N_c \times N_c},
\]
where $A^{road,(c)}$ encodes physical road connectivity, $A^{risk,(c)}$ reflects crash co-occurrence correlations, 
and $A^{poi,(c)}$ models point-of-interest similarity. 
All adjacency matrices are symmetric, i.e., $A_{ij}^{(c)}=A_{ji}^{(c)}$.

\subsection{Problem Formulation}
\paragraph{Prediction target}
For each training sample $k$ and city $c \in \{1, \ldots, C\}$,  
the objective is to predict the accident risk intensity map at the next time step $T{+}1$.  
Formally,
\[
Y_{T+1}^{(k)} = \{Y_{T+1}^{(k,c)}\}_{c=1}^{C}, 
\quad Y_{T+1}^{(k,c)} \in \mathbb{R}^{W_c \times H_c},
\]
where $Y_{T+1}^{(k,c)}$ denotes the two-dimensional risk intensity distribution of city $c$.  
Each element in $Y_{T+1}^{(k,c)}$ corresponds to the expected accident risk level of a spatial grid cell,  
with $W_c$ and $H_c$ representing the north–south and west–east grid resolutions, respectively.

\paragraph{Dataset representation}
The complete multi-city dataset can thus be expressed as
\[
\mathcal{D} =
\Big\{
(\{X^{\mathrm{geo},(k,c)}, X^{\mathrm{sem},(k,c)}, \mathcal{A}^{(c)}\}_{c=1}^{C},
 \{Y_{T+1}^{(k,c)}\}_{c=1}^{C})
\Big\}_{k=1}^{K},
\]
where $X^{\mathrm{geo},(k,c)}$ and $X^{\mathrm{sem},(k,c)}$ represent the geographical and semantic spatio–temporal inputs, respectively,  
and $\mathcal{A}^{(c)}$ denotes the set of heterogeneous adjacency matrices associated with city $c$.  

The learning objective is to train a shared predictive model across all cities that captures transferable spatio–temporal dependencies  
while preserving city-specific heterogeneity through distinct spatial partitions and graph structures $\{M_c, N_c, \mathcal{A}^{(c)}\}$.  
This formulation defines a unified multi-task learning framework that enables consistent and generalizable accident risk forecasting across diverse urban environments.

\begin{definition}[Multi-City Accident Risk Forecasting]
Given historical spatio-temporal sequences from all cities, the objective is to learn a predictive function $f_\theta$, parameterized by $\theta$, that estimates future accident risk maps:
\[
f_\theta:
\big\{ X^{\mathrm{geo},(k,c)}, X^{\mathrm{sem},(k,c)}, \mathcal{A}^{(c)} \big\}_{c=1}^{C}
\longmapsto
\big\{ \hat{Y}_{T+1}^{(k,c)} \big\}_{c=1}^{C},
\quad
\hat{Y}_{T+1}^{(k,c)} \in \mathbb{R}^{W_c \times H_c}.
\]
\end{definition}

This formulation defines a joint multi-task learning problem, where each city represents an individual task with its own spatial and semantic topology.  
The shared model $f_\theta$ captures both transferable patterns across cities and localized variations, forming a unified framework for cross-city accident forecasting and prevention.

\section{Method}
\label{s4}
\begin{figure}[htbp!]
    \centering
    \includegraphics[width=\textwidth]{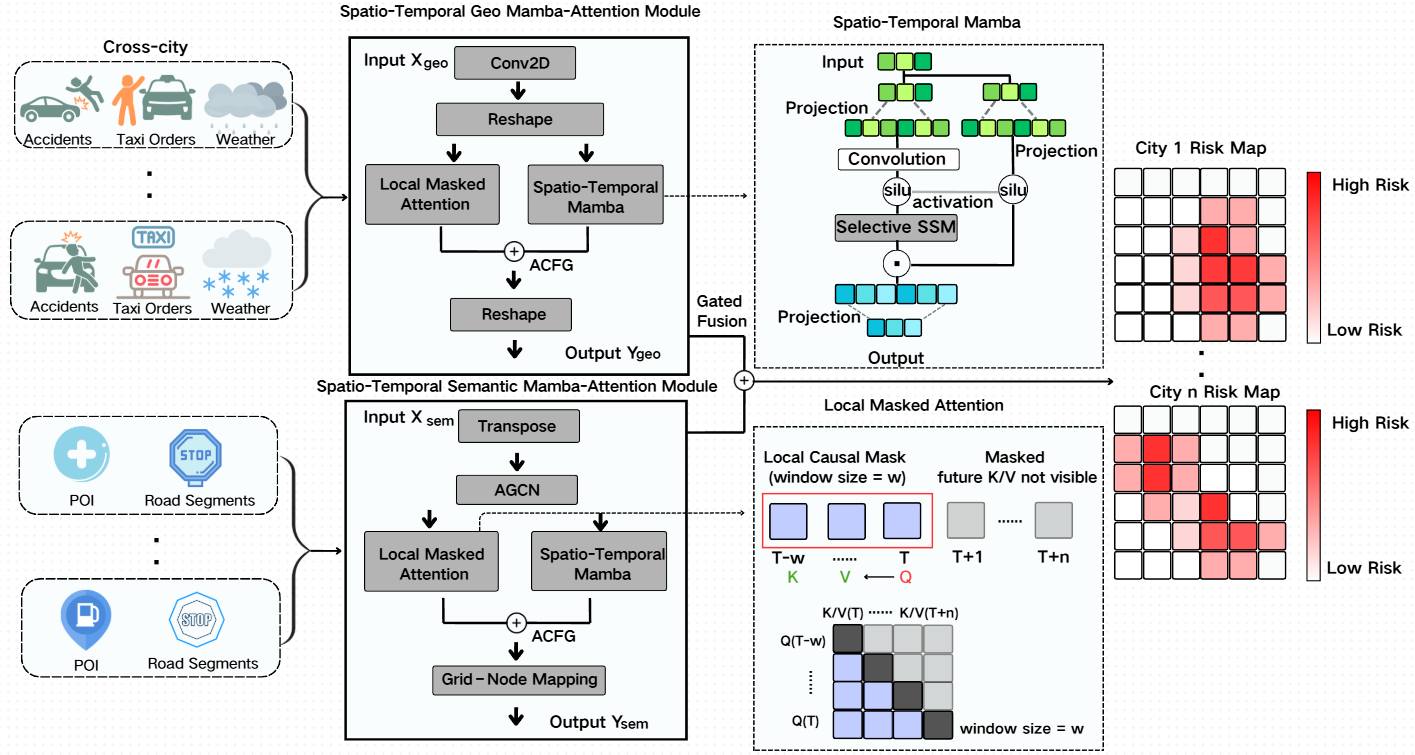} 
    \caption{Structure overview of the proposed \textsf{MLA-STNet} model.}
    \label{structure}
\end{figure}
\subsection{Spatio-Temporal Embedding}

In risk prediction, accident occurrence is influenced by both localized spatial grid patterns 
(e.g., hot spots in specific areas) and relational dependencies across the road network 
(e.g., traffic flows and structural connectivity). 
To effectively capture these two complementary aspects, 
we design the model input as a combination of grid-based (Geographical) 
and node-based (Semantic) spatio–temporal sequences.  Fig.~\ref{structure} illustrates the overall framework of the proposed \textsf{MLA-STNet}, which integrates the Spatio–Temporal Geographical Mamba-Attention (STG-MA) and Spatio–Temporal Semantic Mamba-Attention (STS-MA) modules. These two modules jointly learn localized spatial features and long-range temporal dependencies while preserving city-specific semantic relationships. The dataset is constructed by concatenating multi-city data into a unified representation, enabling joint learning across urban domains, whereas the model outputs are subsequently decomposed into city-specific risk maps for fine-grained accident forecasting.
The dataset is constructed by concatenating data from multiple cities into a unified representation, 
while the outputs are later separated into city-specific risk maps.
For clarity and conciseness, we omit the city index $c$ in the notation 
throughout the input representation and model formulation sections, 
as all variables (e.g., $M$, $N$, $\mathcal{A}$) are defined within a single city context. 
The city index $c$ will be explicitly reintroduced in the prediction stage 
to indicate that the model jointly forecasts risk intensity maps across multiple cities. Specifically, the grid-based temporal observation sequence is denoted as
\[
X^{geo} \in \mathbb{R}^{T \times M \times F_{geo}},
\]
where $T$ is the temporal length, $M = W \cdot H$ is the number of spatial grid cells after concatenation across cities, 
and $F_{geo}$ is the feature dimension for each cell. 
This representation preserves fine-grained temporal–spatial variations in the grid domain. Meanwhile, the node-based temporal observation sequence is denoted as
\[
X^{sem} \in \mathbb{R}^{T \times N \times F_{sem}},
\]
where $N$ is the total number of road network nodes aggregated across all cities, 
and $F_{sem}$ is the node feature dimension. 
To capture heterogeneous spatial relations in this branch, 
we define a set of bidirectional adjacency matrices:
\[
\mathcal{A} = \{A^{road}, A^{risk}, A^{poi}\}, \quad A^{(\cdot)} \in \mathbb{R}^{N \times N},
\]
where $A^{road}$ encodes the physical connectivity of the road network, 
$A^{risk}$ captures correlations derived from historical accident co-occurrences, 
and $A^{poi}$ reflects similarity of point-of-interest distributions.  
In practice, $\mathcal{A}$ has a block-diagonal structure, 
so that relations are only defined within each city rather than across cities.  
All adjacency matrices are symmetric, i.e., $A_{ij} = A_{ji}$, 
thereby encoding bidirectional pairwise relations among nodes.

To construct unified temporal embeddings for both input branches, 
we adopt lightweight two-layer \(1 \times 1\) convolutional structures applied along the feature dimension. 
Formally, for the Geographical branch we have
\begin{equation}
    \mathbf{H}^{geo} = \mathrm{Conv}_{1 \times 1}^{(2)}\big(\mathrm{ReLU}(\mathrm{Conv}_{1 \times 1}^{(1)}(\mathbf{X}^{geo}))\big),
    \quad \mathbf{H}^{geo} \in \mathbb{R}^{T \times M \times D},
\end{equation}
and for the Semantic branch
\begin{equation}
    \mathbf{H}^{sem} = \mathrm{Conv}_{1 \times 1}^{(2)}\big(\mathrm{ReLU}(\mathrm{Conv}_{1 \times 1}^{(1)}(\mathbf{X}^{sem}))\big),
    \quad \mathbf{H}^{sem} \in \mathbb{R}^{T \times N \times D},
\end{equation}
where \(D\) is the embedding dimension. 
The \(1 \times 1\) convolutions act as per-time linear projections that map the raw feature spaces 
(\(F_{geo}\) for the grid input and \(F_{sem}\) for the semantic input) into a shared embedding space of dimension \(D\). 
A ReLU (Rectified Linear Unit) activation is applied between the two convolutional layers, 
which introduces nonlinearity by mapping negative values to zero and thus enhances representation capacity 
while mitigating vanishing gradient issues. 
Formally, this yields two embeddings: 
\(\mathbf{H}^{geo} \in \mathbb{R}^{T \times M \times D}\) for the Geographical branch 
and \(\mathbf{H}^{sem} \in \mathbb{R}^{T \times N \times D}\) for the Semantic branch, 
where \(T\) is the temporal length, \(M=W \cdot H\) is the number of grid cells, and \(N\) is the number of road network nodes.




\subsection{Spatio-Temporal Geographical Mamba-Attention Module (STG-MA)}
\label{geo}

On top of the embedding sequence 
\(\mathbf{H}^{geo}   \in \mathbb{R}^{T \times M \times D}\), 
where \(M=W \cdot H\) is the number of grid cells and each \(\mathbf{h}_t \in \mathbb{R}^{M \times D}\) 
encodes the grid-level latent features at time step \(t\), 
we introduce the Spatio-Temporal Geographical Mamba-Attention Module (STG-MA) to explicitly capture both spatial and temporal dependencies that are critical for risk dynamics.  
This module aims to address a key limitation of existing approaches, which often struggle to capture long-range temporal dependencies and are vulnerable to spurious signals from low-risk regions, leading to unstable predictions under noisy and imbalanced accident records. The underlying motivation is that spatially adjacent regions tend to exhibit correlated risk patterns, while temporal evolution dictates how such risks propagate across time.  

At each time step, we take \(\mathbf{h}_t\) as the spatial snapshot of the entire grid, 
which compactly summarizes the latent features of all \(M\) cells.  
However, representing the grid as a flat index set of size \(M\) discards explicit spatial adjacency information.  
To recover this structure, we reshape \(\mathbf{h}_t\) into a grid representation 
\(\tilde{\mathbf{h}}_t \in \mathbb{R}^{D \times W \times H}\), 
where the embedding dimension \(D\) is treated as the channel dimension and \((W,H)\) encode the original spatial layout of the city map.  
On this reshaped grid, we apply a two-dimensional convolution:
\begin{equation}
    \mathbf{Z}_t = \mathrm{Conv2D}(\tilde{\mathbf{h}}_t), 
    \quad \mathbf{Z}_t \in \mathbb{R}^{D \times W \times H}.
\end{equation}
Here, \(\mathrm{Conv2D}\) uses a kernel size of \(3 \times 3\), stride 1, and padding 1, 
so that the spatial resolution \((W,H)\) is preserved.  
This operation enables the model to capture localized spatial correlations among neighboring cells, 
while still retaining the temporal index \(t\) for subsequent sequential modeling.



Collecting all timesteps gives the sequence 
\(\mathbf{Z} = (\mathbf{Z}_1,\dots,\mathbf{Z}_T) \in \mathbb{R}^{T \times D \times W \times H}\). 
To enable temporal modeling, we rearrange \(\mathbf{Z}\) such that each grid cell 
\((w,h)\) corresponds to a temporal trajectory of length \(T\):
\begin{equation}
    \mathbf{Z}' = \mathrm{ReshapePermute}(\mathbf{Z}), 
    \quad \mathbf{Z}' \in \mathbb{R}^{(W \cdot H) \times T \times D}.
\end{equation}
so that each spatial cell \((w,h)\) is mapped to its own temporal trajectory 
\((\mathbf{z}'_{(w,h),1}, \dots, \mathbf{z}'_{(w,h),T})\).  
This canonical ordering \(((W \cdot H), T, D)\) ensures compatibility with the Mamba and Attention Block, 
which operates along the temporal dimension \(T\).


\subsubsection{Local Masked Attention in STGM (LMA)}
\label{geomask}

The tensor $\mathbf{Z}' \in \mathbb{R}^{(W \cdot H) \times T \times D}$ 
can be regarded as $(W \cdot H)$ parallel sequences, each of length $T$ 
and feature dimension $D$. 
Here, the spatial dimension $(W \cdot H)$ enumerates all grid cells, 
so that each grid cell corresponds to an independent temporal trajectory. 
In practice, the local masked attention operates on sequences of shape 
$[T \times D]$, while the additional $(W \cdot H)$ dimension simply 
represents parallel trajectories that are processed independently. 

To prevent information leakage, a causal mask is applied such that 
at time step $t$, the receptive field is restricted to its preceding $w$ steps. 
Here, $w$ denotes the local receptive window size, which is a tunable 
hyperparameter that controls how many past timesteps each position can attend to. 
This design ensures that future information remains inaccessible while 
enabling the model to capture localized short-term temporal dependencies 
within each spatial trajectory.  

To extract such risk patterns, we employ a masked local multi-head 
attention mechanism. At each step $t$, the representation is projected into 
queries, keys, and values:

\begin{equation}
    \mathbf{Q}^{\text{geo}} = \mathbf{Z}' W_Q^{\text{geo}}, 
    \quad \mathbf{K}^{\text{geo}} = \mathbf{Z}' W_K^{\text{geo}}, 
    \quad \mathbf{V}^{\text{geo}} = \mathbf{Z}' W_V^{\text{geo}},
\end{equation}
where $W_Q^{\text{geo}}, W_K^{\text{geo}}, W_V^{\text{geo}} \in \mathbb{R}^{D \times D}$ are learnable projection matrices applied along the feature dimension.  
Here, $\mathbf{Z}' \in \mathbb{R}^{(W \cdot H) \times T \times D}$ is the reshaped sequence, 
with $(W \cdot H)$ grid cells, temporal length $T$, and hidden size $D$.  

In practice, the hidden size $D$ is split across $m_{\text{heads}}$ heads, 
so that each head has dimension $d = D / m_{\text{heads}}$.  
Thus, per-head queries, keys, and values have shape 
$\mathbb{R}^{(W \cdot H) \times T \times d}$ before concatenation, 
and the equations above are written in the \textit{post-concatenation} form 
(after all heads are merged back to $D$).  

The attention at time step $t$ for grid cell $m$ is restricted to a causal local window of size $w$, i.e., only the past $w$ steps are visible:
\begin{equation}
    \alpha^{\text{geo}}_{m,t,s} = 
    \frac{\exp\!\left(\tfrac{\mathbf{Q}^{\text{geo}}_{m,t} \cdot \mathbf{K}^{\text{geo}}_{m,s}}{\sqrt{d}}\right)}
         {\sum_{u \in \mathcal{M}(t)} \exp\!\left(\tfrac{\mathbf{Q}^{\text{geo}}_{m,t} \cdot \mathbf{K}^{\text{geo}}_{m,u}}{\sqrt{d}}\right)},
    \quad s \in \mathcal{M}(t),
\end{equation}
where 
$m \in [1, W \cdot H]$ indexes spatial grid cells,  
$t \in [1, T]$ is the current timestep,  
$s \in \mathcal{M}(t)=\{t-w,\dots,t\}$ is a valid historical timestep within the local causal window of size $w$,  
and $\mathbf{Q}^{\text{geo}}_{m,t}, \mathbf{K}^{\text{geo}}_{m,s} \in \mathbb{R}^d$ 
are the per-head projected query and key vectors at timesteps $t$ and $s$.  
The normalization ensures that $\alpha^{\text{geo}}_{m,t,s}$ is a valid probability distribution over $\mathcal{M}(t)$.  For early timesteps where $t < w$, the window $\mathcal{M}(t)$ automatically shortens to include only available history.  

This formulation enforces causality and guarantees that each grid cell $m$ learns its own temporal trajectory 
without accessing future information.
After aggregation, the localized representation for grid cell $m$ at time step $t$ is
\begin{equation}
    \mathbf{L}^{\text{geo}}_{m,t} = \sum_{s \in \mathcal{M}(t)} \alpha^{\text{geo}}_{m,t,s} \mathbf{V}^{\text{geo}}_{m,s},
    \quad \mathbf{L}^{\text{geo}}_{m,t} \in \mathbb{R}^{D}.
\end{equation}
Stacking across all timesteps yields $\mathbf{L}^{\text{geo}}_{m} \in \mathbb{R}^{T \times D}$ for a single grid cell,  
and further stacking across the $(W \cdot H)$ spatial positions produces  
$\mathbf{L}^{\text{geo}} \in \mathbb{R}^{(W \cdot H) \times T \times D}$.  
Here, $\mathbf{V}^{\text{geo}}_{m,s}$ is simply the value vector obtained from the standard attention projection,  
and its weighted sum with $\alpha^{\text{geo}}_{m,t,s}$ corresponds to the usual attention score expansion.  
Thus, $\mathbf{L}^{\text{geo}}$ encodes localized short-term temporal dependencies for each spatial trajectory in the grid.

\subsubsection{Spatio-Temporal Mamba in STGM (STM)}
In contrast to the Local Masked Attention (LMA) branch that focuses on short-term temporal dependencies, 
we integrate a Selective State Space Model (Mamba)~\citep{Gu2023MambaLS} to capture long-range temporal dynamics efficiently. 
The input sequence is reshaped as 
$\mathbf{Z}' \in \mathbb{R}^{(W \cdot H) \times T \times D}$, 
where each spatial index $m \in [1, W \cdot H]$
corresponds to a unique grid trajectory 
$\mathbf{z}'_{m,:} \in \mathbb{R}^{T \times D}$ modeled independently along the temporal axis $T$.

For each grid cell $m$, the selective state-space recurrence of Mamba is defined as:
\begin{align}
    \tilde{\mathbf{A}}_{m,t} &= \exp(\Delta t \cdot \mathbf{A}_{\log}) \odot \sigma(\mathbf{W}_a \mathbf{z}'_{m,t}), \\
    \tilde{\mathbf{B}}_{m,t} &= \mathbf{W}_b \mathbf{z}'_{m,t}, \\
    \mathbf{h}^{\prime \text{geo}}_{m,t} &= \tilde{\mathbf{A}}_{m,t} \odot \mathbf{h}^{\prime \text{geo}}_{m,t-1} + \tilde{\mathbf{B}}_{m,t}, \\
    \mathbf{G}^{\text{geo}}_{m,t} &= \mathbf{W}_c \mathbf{h}^{\prime \text{geo}}_{m,t}.
\end{align}
where $\mathbf{z}'_{m,t} \in \mathbb{R}^D$ denotes the input feature of grid cell $m$ at time step $t$, 
and $\mathbf{h}'^{\text{geo}}_{m,t} \in \mathbb{R}^D$ represents its hidden state that evolves along the temporal dimension.  
$\mathbf{A}_{\log} \in \mathbb{R}^{D \times D}$ is a learnable log-spectral transition matrix that governs the continuous-time dynamics, and $\Delta t$ is a trainable scalar controlling the time-step scale.  
$\sigma(\cdot)$ is the selective gating function, adaptively modulating the retention ratio of previous memory based on the current input.  
$\mathbf{W}_a$, $\mathbf{W}_b$, and $\mathbf{W}_c \in \mathbb{R}^{D \times D}$ are learnable projection matrices for gating, input modulation, and output readout, respectively.  

The operator $\odot$ denotes the Hadamard (element-wise) product, 
which gates each latent dimension of $\mathbf{h}^{\prime \text{geo}}_{m,t-1}$ independently, 
enabling fine-grained control over memory retention and update strength across temporal steps.  
The exponential mapping $\exp(\Delta t \cdot \mathbf{A}_{\log})$ guarantees numerical stability 
and allows flexible control over long-range temporal dependencies.  
The hidden state $\mathbf{h}'^{\text{geo}}_{m,0}$ is initialized to zero, i.e., $\mathbf{h}'^{\text{geo}}_{m,0} = \mathbf{0}$.

For each grid cell $m$, this recurrence produces a sequence of outputs $\mathbf{G}^{\text{geo}}_{m,t} \in \mathbb{R}^D$ across all timesteps, 
which are stacked to form $\mathbf{G}^{\text{geo}}_{m} \in \mathbb{R}^{T \times D}$. 
Aggregating over all spatial positions yields the global representation $\mathbf{G}^{\text{geo}} \in \mathbb{R}^{(W \cdot H) \times T \times D}$, 
which captures long-range temporal dependencies while preserving spatial alignment.  
In this way, the Mamba-based STM branch models global, input-adaptive temporal dynamics that complement the local short-term patterns learned by the LMA branch, 
jointly enhancing the expressiveness.




Finally, the local and global features are fused through an  
\textit{Adaptive Channel Fusion Gate (ACFG)}: 
\begin{equation}
    \mathbf{U}^{\text{geo}}_{m,t} = \mathrm{LayerNorm}\!\Big(\mathbf{z}'_{m,t} 
    + W_f^{\text{geo}} \,[\mathbf{L}^{\text{geo}}_{m,t} ; \mathbf{G}^{\text{geo}}_{m,t}]\Big),
    \quad \mathbf{U}^{\text{geo}}_{m,t} \in \mathbb{R}^{D},
\end{equation}
where $m \in [1, W \cdot H]$ indexes spatial grid cells and $t \in [1,T]$ indexes timesteps.  
Here, $\mathbf{U}^{\text{geo}}_{m,t}$ denotes the contextualized representation of grid cell $m$ at time step $t$,  
\([\cdot ; \cdot]\) denotes concatenation, and \(W_f^{\text{geo}} \in \mathbb{R}^{2D \times D}\) projects the fused feature back to hidden size.  
The proposed Adaptive Channel Fusion Gate (ACFG) adaptively balances the contributions of localized features  
(\(\mathbf{L}^{\text{geo}}_{m,t}\)) and long-range features (\(\mathbf{G}^{\text{geo}}_{m,t}\)) on each channel dimension,  
while LayerNorm stabilizes training.  Stacking across all spatial positions yields 
\(\mathbf{U}^{\text{geo}}_{t} \in \mathbb{R}^{(W \cdot H) \times D}\),  
and further across timesteps produces  
\(\mathbf{U}^{\text{geo}} \in \mathbb{R}^{(W \cdot H) \times T \times D}\).  

In practice, we only retain the final timestep representation, 
since the prediction target is the accident risk at the next time step \((T{+}1)\). 
This design ensures that the model relies on the most recent contextualized state of each grid cell, 
while past information has already been integrated into this representation through the LMA and STM branches:
\begin{equation}
    \mathbf{U}^{\text{geo}}_{T} \in \mathbb{R}^{(W \cdot H) \times D}.
\end{equation}
Each row of \(\mathbf{U}^{\text{geo}}_{T}\) corresponds to the contextualized representation of one grid cell at the last observed step.  
By reshaping the spatial dimension back to a 2D grid, we obtain
\begin{equation}
    \mathbf{Y}_{\text{geo}} = \mathrm{Reshape}\!\big(\mathbf{U}^{\text{geo}}_{T}\big),
    \quad \mathbf{Y}_{\text{geo}} \in \mathbb{R}^{D \times W \times H}.
\end{equation}
Thus, \(\mathbf{Y}_{\text{geo}}\) is a spatial feature map where each grid cell contributes a \(D\)-dimensional representation,  
ready for downstream risk forecasting.

\subsection{Spatio-Temporal Semantic Mamba-Attention Module (STS-MA)}

On top of the temporal embeddings 
\(\mathbf{H}^{sem} \in \mathbb{R}^{T \times N \times D}\), 
where $N$ denotes the number of semantic nodes (e.g., road segments), 
we introduce the Semantic Spatio-Temporal Module to refine relational dependencies across heterogeneous semantic graphs 
(road network, risk correlation, and point-of-interest) 
while capturing both local and global temporal dynamics.  
The motivation is that risk propagation is not only governed by spatial proximity, 
but also by heterogeneous semantic relations between nodes, 
and these relations evolve over time.  

Thus, the Semantic Module directly takes as input \((\mathbf{H}^{sem}, \mathcal{A})\), 
where \(\mathcal{A} = \{A^{road}, A^{risk}, A^{poi}\}\) provides heterogeneous adjacency supports, 
and each adjacency matrix $A^{(\cdot)} \in \mathbb{R}^{N \times N}$ encodes one semantic relation type.  

For spatial-semantic modeling, we adopt adaptive graph convolution networks (AGCNs).  
In addition to the fixed heterogeneous supports 
\(\mathcal{A} = \{A^{road}, A^{risk}, A^{poi}\}\), we introduce learnable adaptive supports parameterized as
\begin{equation}
A^{adp} = \mathrm{Softmax}\!\big(\mathrm{ReLU}(E_1 E_2)\big),
\end{equation}
where \(E_1, E_2 \in \mathbb{R}^{N \times r}\) are learnable node embeddings initialized via low-rank SVD decomposition of the corresponding prior adjacency matrices.  
The Softmax is applied row-wise to ensure valid stochastic supports.
 
Thus, the effective support set becomes
\begin{equation}
\widetilde{\mathcal{A}} = \{A^{road}, A^{risk}, A^{poi}, A^{adp}\},
\end{equation}
enabling the model to combine fixed structures with adaptive relations. We initialize the graph convolutional input by transposing the temporal embeddings into a node-centric representation
\begin{equation}
\mathbf{S}^{(0)} = \text{Transpose}(\mathbf{H}^{sem}) \in \mathbb{R}^{N  \times D},
\end{equation}
so that adjacency matrices operate directly on the node dimension at each time step.  
Based on this initialization, we represent the hidden node-centric embeddings as 
\(\mathbf{S}^{(l)} \in \mathbb{R}^{N \times T \times D_l}\) 
at the \(l\)-th AGCN layer, 
where \(N\) is the number of semantic nodes, \(T\) is the temporal horizon,  
and \(D_l\) is the feature dimension at layer \(l\).  
Node features are updated via supports in \(\widetilde{\mathcal{A}}\): 
\begin{equation}
    \mathbf{S}^{(l+1)}_{:,t,:} = \sigma\!\Bigg(\sum_{j=1}^{J} 
    \widehat{\mathbf{A}}_j \,\mathbf{S}^{(l)}_{:,t,:}\, W_j^{(l)}\Bigg),
    \quad t = 1,\dots,T,
\end{equation}
where \(J\) denotes the total number of support matrices (e.g., three fixed semantic graphs plus one adaptive adjacency, so \(J=4\) in our setting).  
Each \(\widehat{\mathbf{A}}_j\) is the normalized support matrix,  
computed as 
\begin{equation}
\widehat{\mathbf{A}}_j = \hat{D}_j^{-\tfrac{1}{2}} A_j \hat{D}_j^{-\tfrac{1}{2}},
\end{equation}
where \(A_j \in \mathbb{R}^{N \times N}\) is the raw adjacency of the \(j\)-th relation  
and \(\hat{D}_j\) is its degree matrix.  
This normalization stabilizes training and ensures scale-invariant message passing.  

Meanwhile, \(W_j^{(l)} \in \mathbb{R}^{D_l \times D_{l+1}}\) is a trainable weight matrix that transforms the feature dimension \(D_l\) into \(D_{l+1}\).  
The non-linear activation \(\sigma(\cdot)\) is then applied element-wise.  
The summation over \(j=1,\dots,J\) means that node features are aggregated from all available relation types,  
and the temporal axis \(T\) is treated independently (broadcasting the same graph operation across all timesteps).  

After stacking \(L\) AGCN layers, the final representation is  
\(\mathbf{S} = \mathbf{S}^{(L)} \in \mathbb{R}^{N \times T \times D}\).  
For clarity, we denote \(\mathbf{S}_t \in \mathbb{R}^{N \times D}\) 
as the slice of \(\mathbf{S}\) at time step \(t\),  
which corresponds to per-node embeddings at time \(t\).  
These per-time-step node embeddings \(\mathbf{S}_t\) are then used as input to the temporal attention module.


\subsubsection{Local Masked Attention in STSM (LMA)}

To capture localized temporal dynamics, 
we apply masked local multi-head attention~\ref{geo}.  
Formally, for each time step $t$,
\begin{align}
    \mathbf{Q}^{\text{sem}}_t &= \mathbf{S}_t W_Q^{\text{sem}}, \quad
    \mathbf{K}^{\text{sem}}_s = \mathbf{S}_s W_K^{\text{sem}}, \quad
    \mathbf{V}^{\text{sem}}_s = \mathbf{S}_s W_V^{\text{sem}}, \\
    \alpha^{\text{sem}}_{n,t,s} &=
    \frac{\exp\!\big((\mathbf{Q}^{\text{sem}}_{n,t} {\mathbf{K}^{\text{sem}}_{n,s}}^\top)/\sqrt{d}\big)}
    {\sum_{s' \in \mathcal{M}(t)} \exp\!\big((\mathbf{Q}^{\text{sem}}_{n,t} {\mathbf{K}^{\text{sem}}_{n,s'}}^\top)/\sqrt{d}\big)}, \\
    \mathbf{L}^{\text{sem}}_{n,t} &= \sum_{s \in \mathcal{M}(t)} 
    \alpha^{\text{sem}}_{n,t,s} \mathbf{V}^{\text{sem}}_{n,s},
    \quad \mathbf{L}^{\text{sem}}_{n,t} \in \mathbb{R}^{D}.
\end{align}
Here, $\mathbf{S}_t \in \mathbb{R}^{N \times D}$ denotes the embeddings of all $N$ nodes at time step $t$, 
and $\mathbf{Q}^{\text{sem}}_t, \mathbf{K}^{\text{sem}}_s, \mathbf{V}^{\text{sem}}_s \in \mathbb{R}^{N \times d}$ 
are the corresponding query, key, and value projections.  
The attention weight $\alpha^{\text{sem}}_{n,t,s}$ is defined per node $n$, 
measuring how node $n$ at time $t$ attends to its own history at context step $s \in \mathcal{M}(t)$.  
Thus, attention operates along the temporal dimension $T$, 
while the node dimension $N$ is treated as parallel independent sequences.  

The resulting features $\mathbf{L}^{\text{sem}} \in \mathbb{R}^{N \times T \times D}$ 
serve as intermediate node-wise temporal representations, 
which are subsequently integrated with the Mamba module for further modeling. To ensure causality, a local causal mask is applied so that, 
at each query step $t$, the receptive field is restricted to the preceding $w$ timesteps.  
This prevents information leakage from the future while enabling the model 
to capture localized short-term dependencies within each node trajectory, 
consistent with the masking strategy described in Section~\ref{geomask}.


\subsubsection{Spatio-Temporal Mamba in STSM (STM)}
In parallel, we employ a Mamba Model~\cite{Gu2023MambaLS} to capture long-range temporal dependencies.  
The input is the semantic node sequence 
$\mathbf{S} \in \mathbb{R}^{N \times T \times D}$, 
where $N$ denotes the number of semantic nodes.  
For each node $n \in [1, N]$, the selective state-space recurrence of Mamba is defined as:
\begin{align}
    \tilde{\mathbf{A}}^{\text{sem}}_{n,t} &= \exp(\Delta t \cdot \mathbf{A}^{\text{sem}}_{\log}) \odot \sigma(\mathbf{W}^{\text{sem}}_a \mathbf{s}_{n,t}), \\
    \tilde{\mathbf{B}}^{\text{sem}}_{n,t} &= \mathbf{W}^{\text{sem}}_b \mathbf{s}_{n,t}, \\
    \mathbf{h}^{\text{sem}}_{n,t} &= \tilde{\mathbf{A}}^{\text{sem}}_{n,t} \odot \mathbf{h}^{\text{sem}}_{n,t-1} + \tilde{\mathbf{B}}^{\text{sem}}_{n,t}, \\
    \mathbf{G}^{\text{sem}}_{n,t} &= \mathbf{W}^{\text{sem}}_c \mathbf{h}^{\text{sem}}_{n,t}.
\end{align}

\noindent
where $\mathbf{s}_{n,t} \in \mathbb{R}^D$ denotes the input embedding of semantic node $n$ at time step $t$, 
and $\mathbf{h}^{\text{sem}}_{n,t} \in \mathbb{R}^D$ represents its hidden state evolving along the temporal dimension.  
$\mathbf{A}^{\text{sem}}_{\log} \in \mathbb{R}^{D \times D}$ is a learnable log-spectral transition matrix governing the continuous-time dynamics, and $\Delta t$ is a trainable scalar controlling the time-step scale.  
$\sigma(\cdot)$ is the selective gating function that adaptively modulates the retention ratio of previous hidden states based on the current input.  
$\mathbf{W}^{\text{sem}}_a$, $\mathbf{W}^{\text{sem}}_b$, and $\mathbf{W}^{\text{sem}}_c \in \mathbb{R}^{D \times D}$ are learnable projection matrices for gating, input modulation, and output readout, respectively.  
This formulation produces the temporal representation $\mathbf{G}^{\text{sem}} \in \mathbb{R}^{N \times T \times D}$, which efficiently captures long-range dependencies in the semantic space and complements the short-term temporal patterns extracted by the LMA branch.

Finally, we fuse the localized and global components via an  
\textit{Adaptive Channel Fusion Gate (ACFG)}:
\begin{equation}
    \mathbf{U}^{\text{sem}}_{n,t} = \mathrm{LayerNorm}\!\Big(\mathbf{s}_{n,t} 
    + W_f^{\text{sem}} \,[\mathbf{L}^{\text{sem}}_{n,t} ; \mathbf{G}^{\text{sem}}_{n,t}]\Big),
    \quad \mathbf{U}^{\text{sem}}_{n,t} \in \mathbb{R}^{D}.
\end{equation}
Here, $\mathbf{U}^{\text{sem}}_{n,t}$ denotes the contextualized representation of node $n$ at step $t$, 
where $W_f^{\text{sem}} \in \mathbb{R}^{2D \times D}$ projects the concatenated short-term 
features $\mathbf{L}^{\text{sem}}_{n,t}$ (from LMA) and long-term features 
$\mathbf{G}^{\text{sem}}_{n,t}$ (from Mamba) back to hidden size.  
Stacking across all nodes and timesteps yields  
$\mathbf{U}^{\text{sem}} \in \mathbb{R}^{N \times T \times D}$.  

In practice, we only retain the last-step representation 
$\mathbf{U}^{\text{sem}}_{T} \in \mathbb{R}^{N \times D}$ 
and project it back to the spatial grid using the predefined  
\textit{grid–node mapping matrix} $\mathbf{M} \in \{0,1\}^{(W \cdot H) \times N}$.  
Since the dataset spans multiple cities, $\mathbf{M}$ has a block-diagonal structure, with each block mapping the nodes of a single city to its own grid cells.
The projection is given by
\begin{equation}
    \mathbf{Y}_{\text{sem}} = \mathrm{Reshape}\!\left(\mathbf{M}\,\mathbf{U}^{\text{sem}}_{T}\right), 
    \quad \mathbf{Y}_{\text{sem}} \in \mathbb{R}^{D \times W \times H}.
\end{equation}
Each grid cell thereby receives a $D$-dimensional embedding aggregated from 
the semantic relations of its associated nodes.  
This ensures semantic outputs are spatially aligned with the grid representation, 
ready for gated fusion with the Geographical branch.

\subsection{Fusion and Output Layer}
For each city $c \in \{1, \dots, C\}$, 
we obtain a geo-spatio-temporal representation 
\(\mathbf{Y}^{\text{geo},(c)} \in \mathbb{R}^{D \times W_c \times H_c}\) 
and a semantic representation 
\(\mathbf{Y}^{\text{sem},(c)} \in \mathbb{R}^{D \times W_c \times H_c}\). 
A gated fusion mechanism is then applied independently to each city to adaptively balance their contributions:
\begin{equation}
    \mathbf{Y}^{\text{fuse},(c)} 
    = \mathrm{GatedFusion}\big(\mathbf{Y}^{\text{geo},(c)}, \mathbf{Y}^{\text{sem},(c)}\big),
    \quad \mathbf{Y}^{\text{fuse},(c)} \in \mathbb{R}^{D \times W_c \times H_c}.
\end{equation}
Here, the fusion is conducted at the \((D \times W_c \times H_c)\) feature-map level within each city's spatial domain, 
rather than at the node level. 
This design ensures that both Geographical and semantic branches are spatially aligned within each grid space, 
avoiding the need for additional node-to-grid mappings and enabling direct, location-wise integration of heterogeneous features. 
The fusion operator \(\mathrm{GatedFusion}\) is implemented as a parameterized sigmoid gate, 
which learns channel-wise weights to adaptively control the relative contributions of 
\(\mathbf{Y}^{\text{geo},(c)}\) and \(\mathbf{Y}^{\text{sem},(c)}\). 
Although the parameters of \(\mathrm{GatedFusion}\) are shared across all cities, the operation is performed independently within each city, 
allowing heterogeneous spatial resolutions while maintaining a unified modeling framework.


Finally, for each city \(c \in \{1, \dots, C\}\), 
the fused representation \(\mathbf{Y}^{\text{fuse},(c)}\) 
is processed by a lightweight output head composed of two successive 
\(1\times1\) convolutional layers with a ReLU activation in between:
\begin{equation}
    \mathbf{Y}_{\text{final}}^{(c)} = 
    \mathrm{Conv}_{1\times1}\!\Big(\mathrm{ReLU}\big(\mathrm{Conv}_{1\times1}(\mathbf{Y}^{\text{fuse},(c)})\big)\Big),
    \quad \mathbf{Y}_{\text{final}}^{(c)} \in \mathbb{R}^{W_c \times H_c}.
\end{equation}
Each \(1\times1\) convolution here functions equivalently to a feature-wise multilayer perceptron (MLP), consisting of a Linear–ReLU–Linear transformation, and is applied independently to each spatial cell within the grid. In other words, this operation transforms feature vectors along the channel dimension without mixing information across spatial locations or temporal steps. The first \(1\times1\) convolution reorganizes and projects the channel features, the ReLU introduces non-linearity, and the second \(1\times1\) convolution maps the fused feature map into a single-channel spatial prediction.

Consequently, the model produces one complete accident risk intensity map
\(\mathbf{Y}_{\text{final}}^{(c)}\) for each city \(c \in \{1,\dots,C\}\), 
where each \(\mathbf{Y}_{final}^{(c)} \in \mathbb{R}^{W_c \times H_c}\) 
represents the predicted spatial distribution of accident risk 
over the entire grid of city \(c\) for the next time step. The detailed procedure of the proposed method is summarized in Algorithm \ref{Algorithm}.

\begin{algorithm}[htbp!]
\caption{MLA-STNet: Mamba Local-
Attention Spatial–Temporal Net- work}
\textbf{Input:} $\{X^{geo,(k,c)}, X^{sem,(k,c)}, \mathcal{A}^{(c)}\}_{c=1}^C$, 
where $X^{geo,(k,c)} \in \mathbb{R}^{T \times M_c \times F_{geo}}$ denotes the grid-based spatio–temporal features of city $c$ in sample $k$, 
$X^{sem,(k,c)} \in \mathbb{R}^{T \times N_c \times F_{sem}}$ represents the node-based semantic features on the road network of city $c$, 
and $\mathcal{A}^{(c)} = \{A^{road,(c)}, A^{risk,(c)}, A^{poi,(c)}\}$ contains the multi-graph adjacency matrices describing road topology, crash co-occurrence, and point-of-interest similarity, respectively. \\[0.3em]
\textbf{Output:} $\{Y_{T+1}^{(k,c)}\}_{c=1}^C$, 
where $Y_{T+1}^{(k,c)} \in \mathbb{R}^{W_c \times H_c}$ denotes the predicted risk map for city $c$ 
\begin{algorithmic}[1]
\State $\mathbf{H}^{geo,(k,c)} \gets \mathrm{Embed1x1}(X^{geo,(k,c)})$
\State $\mathbf{H}^{sem,(k,c)} \gets \mathrm{Embed1x1}(X^{sem,(k,c)})$
\State $\mathbf{Y}_{geo}^{(k,c)} \gets \mathrm{STG\_MA}(\mathbf{H}^{geo,(k,c)})$
\Statex \hspace{1em} $\mathbf{Z}^{geo,(k,c)} \gets \mathrm{Conv2D}(\mathbf{H}^{geo,(k,c)})$
\Statex \hspace{1em} $\mathbf{L}^{geo,(k,c)} \gets \mathrm{LocalMambaAttention}(\mathbf{Z}^{geo,(k,c)})$
\Statex \hspace{1em} $\mathbf{G}^{geo,(k,c)} \gets \mathrm{SpatioTemporalMemory}(\mathbf{Z}^{geo,(k,c)})$
\Statex \hspace{1em} $\mathbf{U}^{geo,(k,c)} \gets \mathrm{AdaptiveCrossFusion}(\mathbf{Z}^{geo,(k,c)}, \mathbf{L}^{geo,(k,c)}, \mathbf{G}^{geo,(k,c)})$
\Statex \hspace{1em} $\mathbf{Y}_{geo}^{(k,c)} \gets \mathrm{ReshapeToGrid}(\mathbf{U}^{geo,(k,c)})$

\State $\mathbf{Y}_{sem}^{(k,c)} \gets \mathrm{STS\_MA}(\mathbf{H}^{sem,(k,c)}, \mathcal{A}^{(c)})$
\Statex \hspace{1em} $\mathbf{S}^{(k,c)} \gets \mathrm{AdaptiveGraphConv}(\mathbf{H}^{sem,(k,c)}, \mathcal{A}^{(c)})$
\Statex \hspace{1em} $\mathbf{L}^{sem,(k,c)} \gets \mathrm{LocalMambaAttention}(\mathbf{S}^{(k,c)})$
\Statex \hspace{1em} $\mathbf{G}^{sem,(k,c)} \gets \mathrm{SpatioTemporalMemory}(\mathbf{S}^{(k,c)})$
\Statex \hspace{1em} $\mathbf{U}^{sem,(k,c)} \gets \mathrm{AdaptiveCrossFusion}(\mathbf{S}^{(k,c)}, \mathbf{L}^{sem,(k,c)}, \mathbf{G}^{sem,(k,c)})$
\Statex \hspace{1em} $\mathbf{Y}_{sem}^{(k,c)} \gets \mathrm{GridProject}(\mathbf{U}^{sem,(k,c)})$

\State $\mathbf{Y}^{fuse,(k,c)} \gets \mathrm{GatedFusion}(\mathbf{Y}_{geo}^{(k,c)}, \mathbf{Y}_{sem}^{(k,c)})$
\State $Y_{T+1}^{(k,c)} \gets \mathrm{OutputHead}(\mathbf{Y}^{fuse,(k,c)})$

\State \Return $\{Y_{T+1}^{(k,c)}\}_{c=1}^C$
\end{algorithmic}\label{Algorithm}
\end{algorithm}

\section{Experiment}
\label{s5}
\subsection{Experimental Setup}

In this study, the urban areas of both New York City (New York City (NYC)) and Chicago were divided into a $20 \times 20$ grid, where each cell covered a spatial extent of $2\,\text{km} \times 2\,\text{km}$. Due to the absence of road networks in certain regions, only 243 valid cells were retained in the New York City (NYC) dataset and 197 in the Chicago dataset.  

During training, model performance was evaluated on the validation set at the end of each epoch. Whenever the validation loss decreased, the corresponding model state was saved. To reduce the risk of overfitting, an early stopping strategy was applied, which terminated training if no improvement was observed over 10 consecutive epochs. In this study, the temporal configuration was designed as one target slot ($Q=1$), six short-term slots ($O=6$), and six long-term slots ($P=6$). This configuration enables the model to capture richer temporal dependencies by incorporating both extended short-term fluctuations and longer-term periodic patterns, which are crucial for improving forecasting accuracy in complex urban environments.  

To comprehensively assess model performance, we adopt three widely used evaluation metrics:  
(1) Root Mean Square Error (RMSE), which quantifies the overall deviation between predicted and observed accident risk intensities; (2) Recall (\%), which measures the proportion of true accident hotspots correctly identified by the model; and (3) Mean Average Precision (MAP), which evaluates the ranking quality of predicted hotspot risk scores. Together, these metrics reflect both numerical accuracy and the model’s ability to capture high-risk areas critical for practical deployment, with detailed computation formulas and notation definitions provided in Table~\ref{eval}.

All experiments were implemented in PyTorch. Our proposed \textsf{MLA-STNet} (Mamba-Local Attention Spatio-Temporal Network), along with the baseline models, was trained and evaluated on a server equipped with a NVIDIA Tesla V100S-PCIE-32GB GPU and 24 CPU cores. We highlight the best results in \textbf{bold}.

\begin{table}[htbp!]
\centering
\footnotesize
\caption{Evaluation metrics used for model assessment.}
\label{eval}
\resizebox{\textwidth}{!}{
\begin{tabular}{|m{3cm}|>{\centering\arraybackslash}m{6cm}|m{6cm}|}
\hline
\textbf{Metric} & \textbf{Formula} & \textbf{Interpretation} \\
\hline
Root Mean Square Error (RMSE) 
& 
$\displaystyle \mathrm{RMSE} = \sqrt{\tfrac{1}{n}\sum_{i=1}^{n}(y_i - \hat{y}_i)^2}$ 
& $n$ is the number of samples, $y_i$ is the ground-truth accident risk intensity, and $\hat{y}_i$ is the predicted value. RMSE measures the overall deviation, penalizing larger errors quadratically. \\
\hline
Recall (\%) 
& 
$\displaystyle \mathrm{Recall} = \tfrac{TP}{TP+FN} \times 100\%$ 
& $TP$ (true positives) is the number of correctly detected hotspots, and $FN$ (false negatives) is the number of actual hotspots missed by the model. Recall measures the proportion of true hotspots successfully identified. \\
\hline
Mean Average Precision (MAP) 
& 
$\displaystyle \mathrm{MAP} = \tfrac{1}{Q}\sum_{q=1}^{Q} \tfrac{1}{m_q}\sum_{k=1}^{m_q} P_q(k)$ 
& $Q$ is the number of queries, $m_q$ is the number of relevant hotspots for query $q$, and $P_q(k)$ is the precision at the $k$-th relevant hotspot in the ranked prediction list. MAP evaluates ranking quality, rewarding models that rank true hotspots higher. \\
\hline
\end{tabular}
}
\end{table}

\subsection{Experimental Results and Analysis}

\begin{figure}[htbp!]
    \centering
    \includegraphics[width=\textwidth]{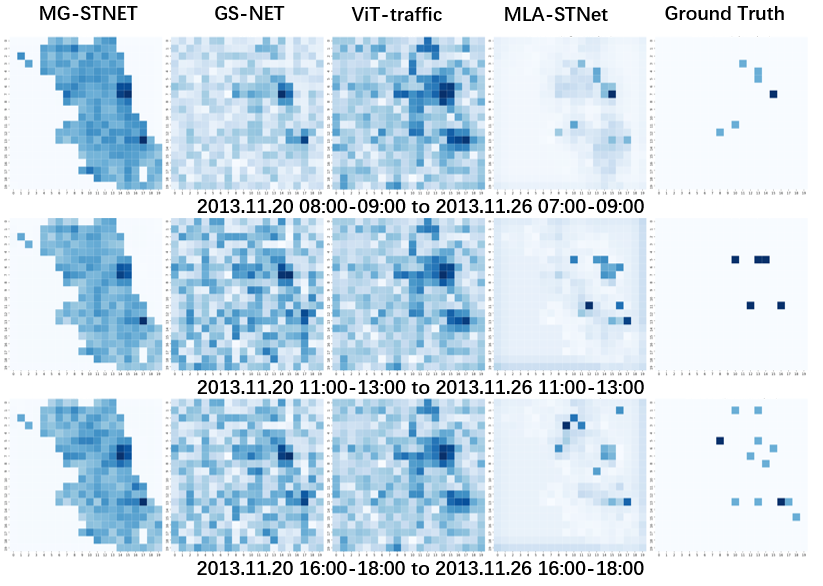}
    \caption{Comparison of accident risk prediction among MG-STNet, GSNet, ViT-Traffic, and \textsf{MLA-STNet} against the Ground Truth on the New York City (NYC) dataset across three representative time periods (2013.11.20–2013.11.26) at 08:00–09:00, 13:00–14:00, and 17:00–18:00.}
    \label{fig:all}
\end{figure}

Fig.~\ref{fig:all} provides a qualitative comparison of spatial accident risk maps predicted by different models on the New York City (NYC) dataset. Existing baselines illustrate clear limitations: GS-Net and ViT-traffic spread responses diffusely across the map, activating many irrelevant regions and failing to reflect the sparse yet spatially clustered nature of real accident patterns. MG-STNet improves stability but still produces overly broad hotspot areas, which blur the boundaries of true accident clusters and exaggerate risk across surrounding regions. These observations highlight two long-standing gaps: (1) the inability of prior models to suppress noisy activations in low-risk areas, and (2) their weakness in faithfully capturing the localized clustering of accident hotspots.

In contrast, \textsf{MLA-STNet} produces risk maps that are both sharper and more consistent with the ground truth. As shown in multiple periods, \textsf{MLA-STNet} concentrates predictive intensity on compact clusters that align with real accident hotspots, while keeping background regions inactive. This balance demonstrates that the model not only suppresses spurious noise but also preserves the intrinsic spatial aggregation of high-risk zones. Such improvement directly stems from the complementary roles of STG-MA and STS-MA: the former enhances temporal stability and filters low-risk noise, while the latter leverages heterogeneous semantic context to refine spatial clustering across different urban environments. Together, these mechanisms bridge the gap between diffuse or over-smoothed baseline predictions and the clustered, imbalanced structure of true accident records.

\begin{table*}[t]
\centering
\caption{Performance comparison of single-task on Chicago and New York City datasets. Best results are in bold.}
\label{tab:performance}

\begin{subtable}{\textwidth}
\centering
\caption{Chicago dataset - Single-task performance}
\resizebox{0.85\textwidth}{!}{
\begin{tabular}{l|ccc|ccc}
\hline
\multirow{2}{*}{Model (single-task)} & \multicolumn{3}{c|}{All-day Performance} & \multicolumn{3}{c}{High-frequency Period} \\
\cline{2-7}
 & RMSE & Recall (\%) & MAP & RMSE & Recall (\%) & MAP \\
\hline
GSNet & 10.63$\pm$0.15 & 21.73$\pm$0.12 & 0.088$\pm$0.009 & 9.10$\pm$0.12 & 22.88$\pm$0.13 & 0.110$\pm$0.010 \\
TWCCnet & 11.13$\pm$0.14 & 20.63$\pm$0.11 & 0.074$\pm$0.010 & 9.50$\pm$0.17 & 21.19$\pm$0.12 & 0.094$\pm$0.012 \\
ViT-Traffic & 9.41$\pm$0.16 & 19.44$\pm$0.10 & 0.088$\pm$0.008 & 6.89$\pm$0.12 & 20.71$\pm$0.11 & 0.106$\pm$0.010 \\
MG-STNet & 9.12$\pm$0.13 & 22.12$\pm$0.09 & 0.099$\pm$0.011 & \textbf{6.45$\pm$0.12} & \textbf{23.18$\pm$0.09} & \textbf{0.121$\pm$0.013} \\
\rowcolor{gray!20}\textbf{\textsf{MLA-STNet}} & \textbf{8.91$\pm$0.13} & \textbf{23.01$\pm$0.08} & \textbf{0.101$\pm$0.010} & 6.61$\pm$0.14 & 22.85$\pm$0.09 & 0.115$\pm$0.013 \\
\hline
\end{tabular}
}
\end{subtable}

\vspace{0.3cm}

\begin{subtable}{\textwidth}
\centering
\caption{New York City dataset - Single-task performance}
\resizebox{0.85\textwidth}{!}{
\begin{tabular}{l|ccc|ccc}
\hline
\multirow{2}{*}{Model (single-task)} & \multicolumn{3}{c|}{All-day Performance} & \multicolumn{3}{c}{High-frequency Period} \\
\cline{2-7}
 & RMSE & Recall (\%) & MAP & RMSE & Recall (\%) & MAP \\
\hline
GSNet & 7.67$\pm$0.18 & 33.41$\pm$0.10 & 0.186$\pm$0.012 & 6.84$\pm$0.13 & 34.50$\pm$0.11 & 0.180$\pm$0.010 \\
TWCCnet & 7.87$\pm$0.13 & 32.41$\pm$0.10 & 0.156$\pm$0.010 & 6.94$\pm$0.19 & 33.50$\pm$0.11 & 0.160$\pm$0.013 \\
ViT-Traffic & 7.20$\pm$0.18 & 32.54$\pm$0.09 & 0.184$\pm$0.012 & 6.48$\pm$0.13 & 32.82$\pm$0.10 & 0.176$\pm$0.009 \\
MG-STNet & 7.13$\pm$0.14 & 34.76$\pm$0.10 & 0.193$\pm$0.010 & 6.45$\pm$0.15 & \textbf{35.58$\pm$0.09} & \textbf{0.187$\pm$0.012} \\
\rowcolor{gray!20}\textbf{\textsf{MLA-STNet}} & \textbf{7.02$\pm$0.12} & \textbf{34.77$\pm$0.09} & \textbf{0.193$\pm$0.012} & \textbf{6.45$\pm$0.11} & 34.90$\pm$0.09 & 0.182$\pm$0.010 \\
\hline
\end{tabular}
}
\end{subtable}
\end{table*}

\begin{table*}[htbp!]
\centering
\caption{Performance comparison of multi-task models on Chicago and New York City datasets.}
\label{tab:performance_combined}

\begin{subtable}{\textwidth}
\centering
\caption{Chicago dataset - Multi-task model performance}
\resizebox{0.85\textwidth}{!}{
\begin{tabular}{l|ccc|ccc}
\hline
\multirow{2}{*}{Model (multi-task)} & \multicolumn{3}{c|}{All-day Performance} & \multicolumn{3}{c}{High-frequency Period} \\
\cline{2-7}
 & RMSE & Recall (\%) & MAP & RMSE & Recall (\%) & MAP \\
\hline
GSNet & 12.65$\pm$0.22 & 18.75$\pm$0.17 & 0.076$\pm$0.012 & 10.92$\pm$0.19 & 19.12$\pm$0.16 & 0.089$\pm$0.011 \\
ViT-Traffic & 11.42$\pm$0.20 & 17.21$\pm$0.16 & 0.075$\pm$0.011 & 8.65$\pm$0.18 & 18.02$\pm$0.15 & 0.087$\pm$0.010 \\
\rowcolor{gray!20}\textbf{\textsf{MLA-STNet}} & \textbf{8.59$\pm$0.11} & \textbf{23.94$\pm$0.09} & \textbf{0.104$\pm$0.012} & \textbf{6.45$\pm$0.10} & \textbf{23.13$\pm$0.08} & \textbf{0.116$\pm$0.011} \\
\hline
\end{tabular}
}
\end{subtable}

\vspace{0.3cm}

\begin{subtable}{\textwidth}
\centering
\caption{New York City dataset - Multi-task model performance}
\resizebox{0.85\textwidth}{!}{
\begin{tabular}{l|ccc|ccc}
\hline
\multirow{2}{*}{Model (multi-task)} & \multicolumn{3}{c|}{All-day Performance} & \multicolumn{3}{c}{High-frequency Period} \\
\cline{2-7}
 & RMSE & Recall (\%) & MAP & RMSE & Recall (\%) & MAP \\
\hline
GSNet & 9.81$\pm$0.24 & 30.28$\pm$0.15 & 0.168$\pm$0.013 & 8.54$\pm$0.21 & 30.97$\pm$0.14 & 0.162$\pm$0.012 \\
ViT-Traffic & 9.23$\pm$0.22 & 29.74$\pm$0.13 & 0.167$\pm$0.012 & 8.06$\pm$0.19 & 29.89$\pm$0.12 & 0.160$\pm$0.011 \\
\rowcolor{gray!20}\textbf{\textsf{MLA-STNet}} & \textbf{6.88$\pm$0.12} & \textbf{34.88$\pm$0.09} & \textbf{0.196$\pm$0.013} & \textbf{6.42$\pm$0.10} & \textbf{35.22$\pm$0.10} & \textbf{0.183$\pm$0.012} \\
\hline
\end{tabular}
}
\end{subtable}
\end{table*}

\paragraph{Performance Evaluation}
Table~\ref{tab:performance} and Table~\ref{tab:performance_combined} summarize the results of all compared models under both single-task and multi-task settings on the Chicago and New York City (NYC) datasets.

In the single-task setting as in Table~\ref{tab:performance}, where models are trained separately on each city, \textsf{MLA-STNet} shows strong and balanced performance across most evaluation metrics. On the Chicago dataset, it achieves an RMSE of 8.91 with Recall of 23.01\% and MAP of 0.101, performing better than GSNet, TWCCNet, and ViT-Traffic, and closely matching the performance of MG-STNet. On the New York City (NYC) dataset, \textsf{MLA-STNet} achieves an RMSE of 7.02, Recall of 34.77\%, and MAP of 0.193, indicating its ability to capture spatial and temporal dependencies effectively even under sparse and noisy accident data. During high-frequency accident periods, the model maintains stable results, reaching Recall around 34.9\% and MAP near 0.182.

In the multi-task setting shown in Table~\ref{tab:performance_combined}, \textsf{MLA-STNet} employs a shared-parameter architecture in which all cities share a unified set of model parameters, while each city independently learns its own feature representations. This design allows the model to leverage the advantages of joint optimization without forcing feature homogenization across cities. As shown in Table~\ref{tab:performance_combined}, this approach leads to consistent improvements compared to the single-task configuration. On the Chicago dataset, RMSE decreases to 8.59 and Recall rises to 23.94\%. On the New York City (NYC) dataset, RMSE drops to 6.88, with Recall reaching 34.88\% and MAP 0.196. These results demonstrate that sharing parameters while allowing each city to learn its own representations enables \textsf{MLA-STNet} to generalize better across heterogeneous urban mobility environments. Under high-frequency accident periods, the model continues to perform robustly, maintaining Recall above 35\% and MAP around 0.18.

The results provide three main insights. First, \textsf{MLA-STNet} achieves strong and stable performance in the single-task setting, confirming its capability in modeling complex spatio-temporal dependencies. Second, the multi-task framework, based on shared parameters but independent feature learning, effectively improves generalization and stability across different metropolitan areas. Third, the robustness of \textsf{MLA-STNet} under high-demand conditions highlights its potential for large-scale, real-world deployment. Overall, \textsf{MLA-STNet} presents a scalable and reliable framework for cross-city urban accident forecasting.

\subsection{Ablation Study}

\paragraph{Spatio-Temporal Geographical Mamba-Attention (STG-MA)}

The primary goal of the Spatio-Temporal Geographical Mamba-Attention (STG-MA) module is to enhance the model’s ability to capture long-range temporal dependencies and localized spatial patterns in highly irregular accident data. By integrating Mamba sequence kernels with localized masked attention, STG-MA is specifically designed to model sequential dynamics over extended horizons while selectively filtering irrelevant or spurious signals.

This module addresses several intrinsic challenges of accident records. First, the data are highly clustered and sparse, with risk concentrated in a limited number of hotspots while most regions exhibit little to no activity. Second, accident records are cyclical and noisy, influenced by periodic urban mobility patterns, weather variations, and reporting inconsistencies. Third, the imbalance between high-risk and low-risk regions creates difficulties for conventional models, which often overfit dense clusters and underperform in rare-event areas. By combining the efficiency of Mamba kernels for sequence modeling with the adaptability of localized masked attention, STG-MA effectively mitigates these challenges, leading to more stable and robust forecasting under noisy, imbalanced, and heterogeneous accident data distributions.

\begin{figure}[htbp!]
    \centering
    \includegraphics[width=\textwidth]{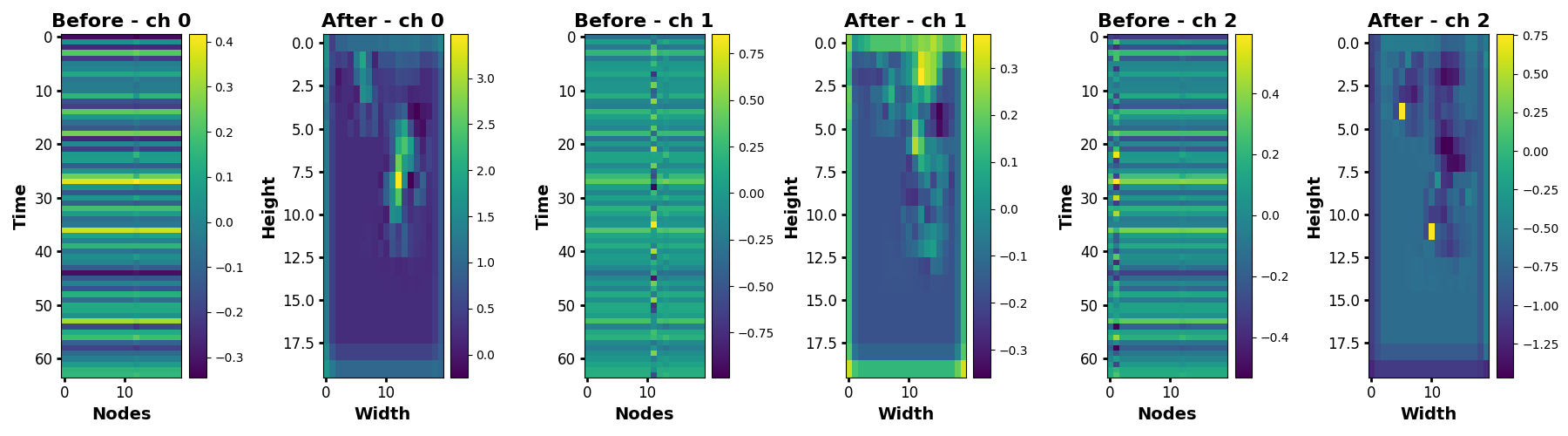} 
    \caption{Feature transformation by STG-MA on the New York City (NYC) dataset.}

    \label{STG-MA Feature}
\end{figure}

\begin{figure}[htbp!]
    \centering
    \begin{minipage}{0.48\textwidth}
        \centering
        \includegraphics[width=\linewidth]{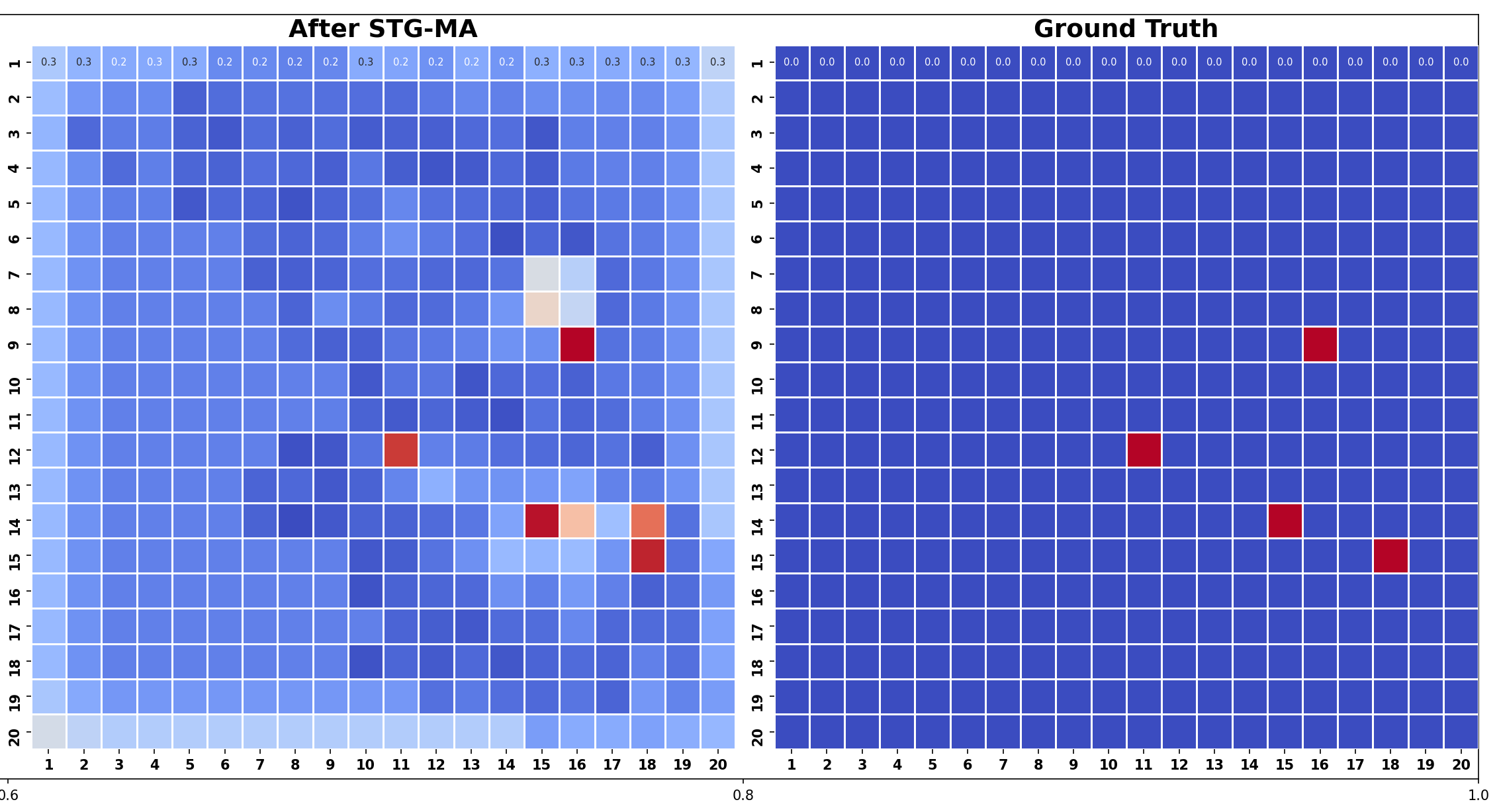}
    \end{minipage}
    \hfill
    \begin{minipage}{0.48\textwidth}
        \centering
        \includegraphics[width=\linewidth]{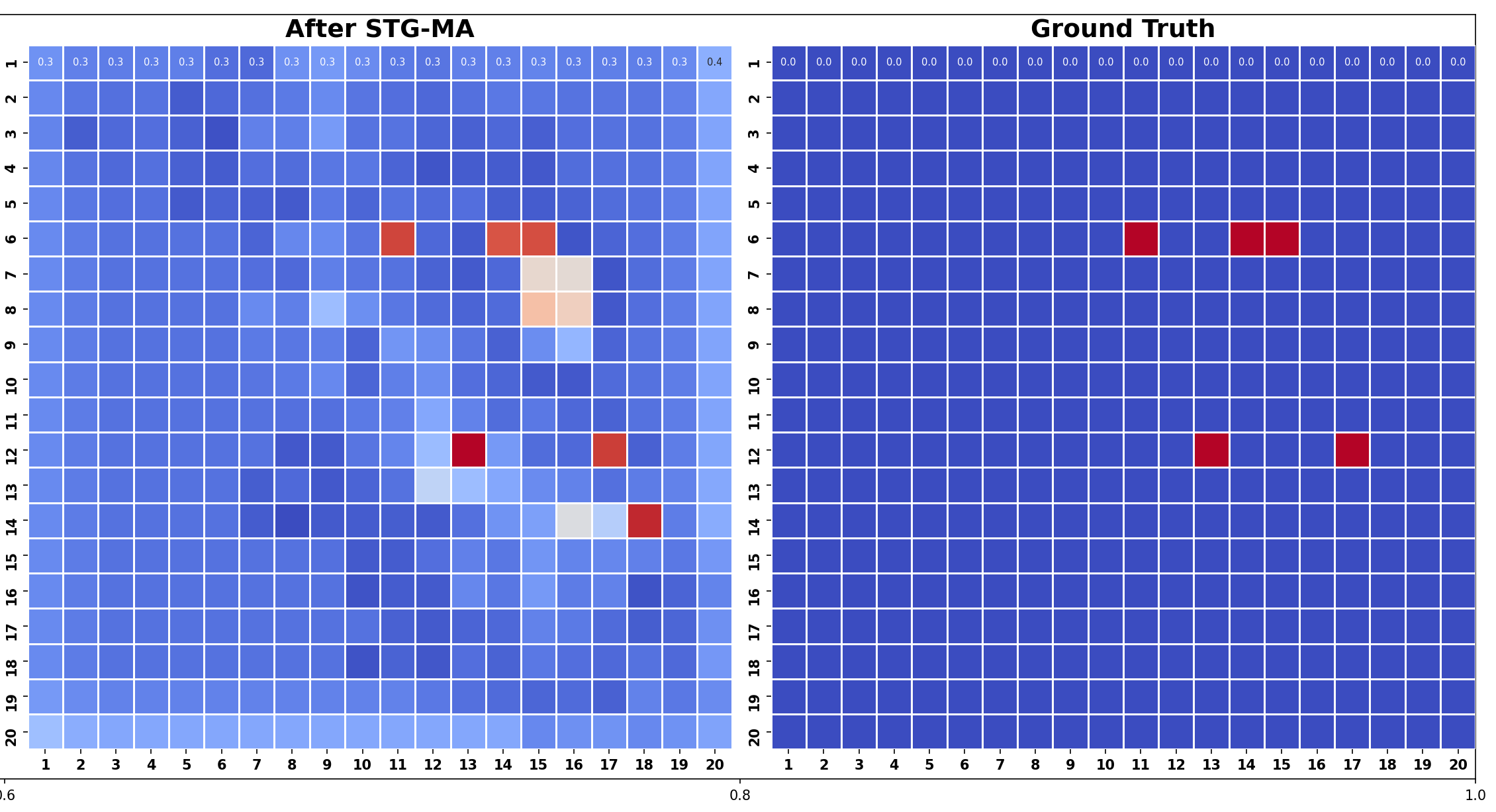}
    \end{minipage}
    \caption{After STG-MA Predictions vs. Ground Truth under Imbalanced Accident Records on the New York City (NYC) dataset.}
    \label{STG-MA VS Truth}
\end{figure}

\paragraph{Effectiveness of STG-MA}
To validate the capability of the proposed Spatio-Temporal Geographical Mamba-Attention (STG-MA) module, we provide feature visualizations before and after its application on the New York City (NYC) dataset. As shown in Fig.~\ref{STG-MA Feature}, the Before representations exhibit repetitive stripe-like activations across time and nodes, indicating that the model fails to capture meaningful temporal variations and spatial heterogeneity. In contrast, the \emph{After} representations demonstrate two critical improvements: (i) activations become temporally structured rather than uniform, showing that long-range dependencies are emphasized while redundant fluctuations are suppressed; and (ii) activations concentrate on localized spatial regions, reflecting the selective filtering mechanism of masked attention that isolates informative accident-prone areas from irrelevant background noise. Together, these changes confirm that STG-MA enhances both temporal discriminability and spatial locality in the learned features.

Another key challenge of accident records lies in the severe imbalance between high-risk and low-risk regions. Conventional models often struggle in such settings, as they may exaggerate dense clusters while overlooking minority occurrences, leading to widespread false alarms or missed detections. To examine whether STG-MA alleviates this imbalance, we compare predicted risk maps with ground truth distributions on the New York City (NYC) dataset, as illustrated in Fig.~\ref{STG-MA VS Truth}. The visualizations reveal that after applying STG-MA, the model more accurately highlights true accident hotspots while simultaneously suppressing spurious activations in low-risk regions. Instead of distributing noisy responses across the spatial domain, STG-MA concentrates predictive intensity on genuine risk areas and maintains stability in sparse backgrounds. This selective enhancement demonstrates the module’s ability to mitigate class imbalance by retaining sensitivity to rare but critical events without overwhelming the predictions with false positives.

\paragraph{Spatio-Temporal Semantic Mamba-Attention (STS-MA)}

The primary goal of the Spatio-Temporal Semantic Mamba-Attention (STS-MA) module is to address the heterogeneity of multi-city accident datasets by jointly modeling shared semantic structures and city-specific variations within a unified framework. By constructing heterogeneous semantic graphs that incorporate roadway topology, POI distributions, urban mobility patterns, meteorological conditions, and historical accident records, STS-MA is designed to capture diverse factors that shape accident risks in different urban environments. 

This module tackles several critical challenges of multi-city accident prediction. First, accident records are inherently heterogeneous across cities, as differences in road networks, functional zoning, and mobility patterns create distinct spatial and semantic structures. Second, the contribution of various semantic factors is highly city-dependent: for example, POI density may strongly influence accidents in commercial centers, whereas weather fluctuations may play a larger role in coastal cities. Third, building entirely separate models for each city prevents the discovery of common patterns and hinders the ability to exploit cross-city information. By combining the temporal modeling capacity of Mamba kernels with semantic attention over heterogeneous graphs, STS-MA unifies these diverse signals into a single representational space. This design allows the model to emphasize shared accident-inducing structures while preserving local nuances, thereby enhancing both the accuracy and robustness of multi-city accident forecasting.

\paragraph{Effectiveness of STS-MA}

\begin{figure}[htbp!]
    \centering
    \includegraphics[width=\textwidth]{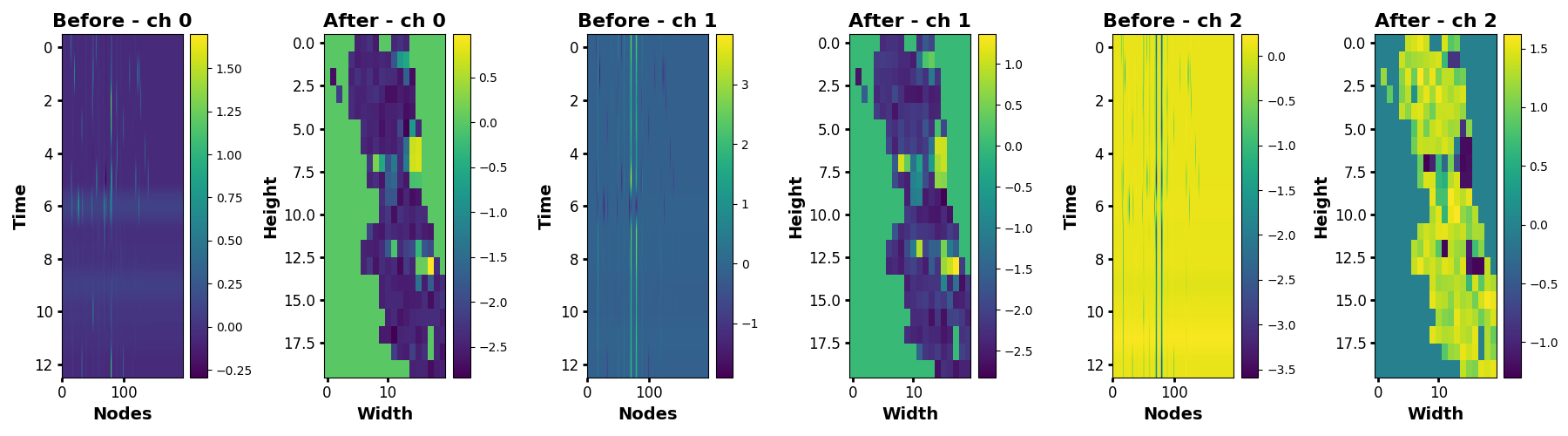} 
    \caption{Feature transformation by STS-MA on the New York City (NYC) dataset}

    \label{structure1}
\end{figure}

To further validate the effectiveness of the proposed Spatio-Temporal Semantic Mamba-Attention (STS-MA) module, we provide qualitative visualizations of its feature transformations on the New York City (NYC) dataset. As shown in Fig.~\ref{structure1}, the semantic node representations before applying STS-MA are noisy and scattered, with little structural coherence. After passing through STS-MA, the representations evolve into more compact and structured clusters, indicating that the module effectively disentangles noisy signals and captures meaningful semantic dependencies across heterogeneous urban environments.

\begin{figure}[htbp!]
    \centering
    \includegraphics[width=\textwidth]{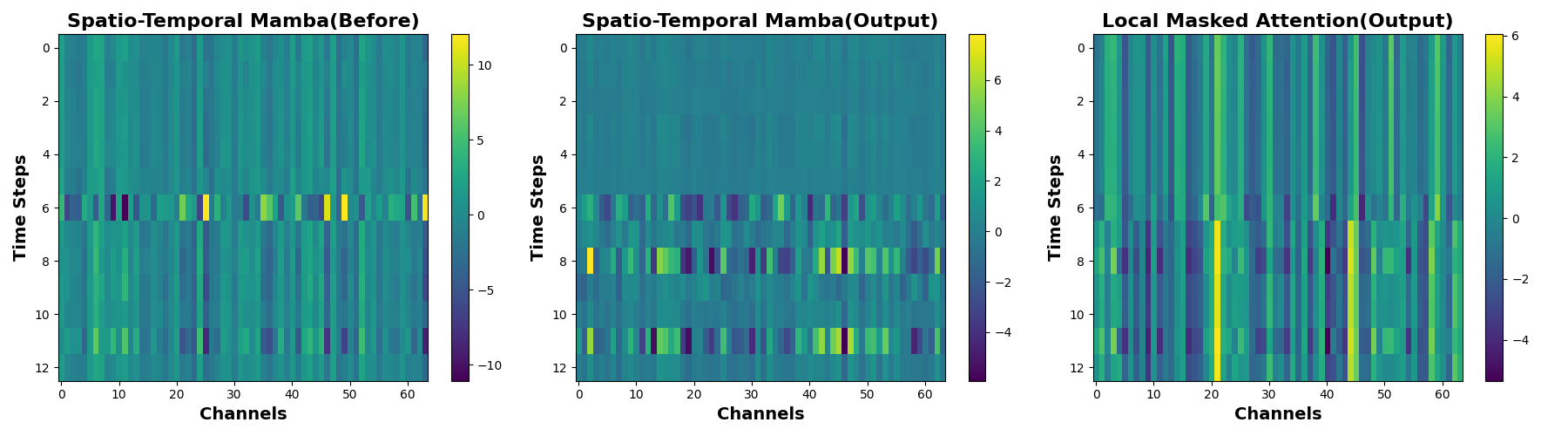} 
    \caption{Visualization of representations in ST-Mamba and Local Masked Attention}
    \label{Mam and Local}
\end{figure}

In addition, Fig.~\ref{Mam and Local} illustrates the transformation of feature representations within the Spatio–Temporal Mamba and Local Masked Attention modules, highlighting the progressive refinement of temporal–channel patterns from the initial input to the final outputs on the New York City (NYC) dataset. It provides an in-depth view of how STS-MA integrates Spatio-Temporal Mamba and Local Masked Attention. The visualization highlights three stages: the input of Spatio-Temporal Mamba (Before), its transformed output, and the output of Local Masked Attention. The results demonstrate that Spatio-Temporal Mamba successfully extracts long-range temporal dependencies across irregular sequences, while Local Masked Attention further refines localized spatial interactions by filtering irrelevant signals. The complementary effects of these two components explain why STS-MA consistently improves model robustness and forecasting accuracy under noisy, imbalanced, and heterogeneous urban mobility distributions.

\begin{figure}[htbp!]
    \centering
    \begin{subfigure}{0.32\textwidth}
        \includegraphics[width=\linewidth]{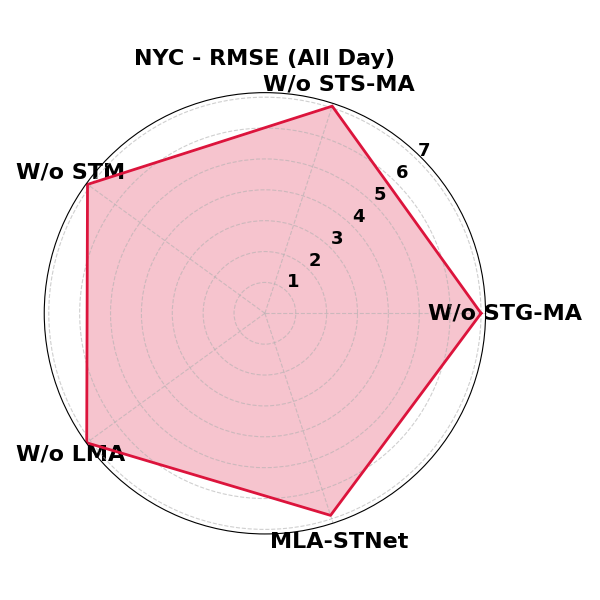}
        \caption{RMSE comparison}
    \end{subfigure}
    \hfill
    \begin{subfigure}{0.32\textwidth}
        \includegraphics[width=\linewidth]{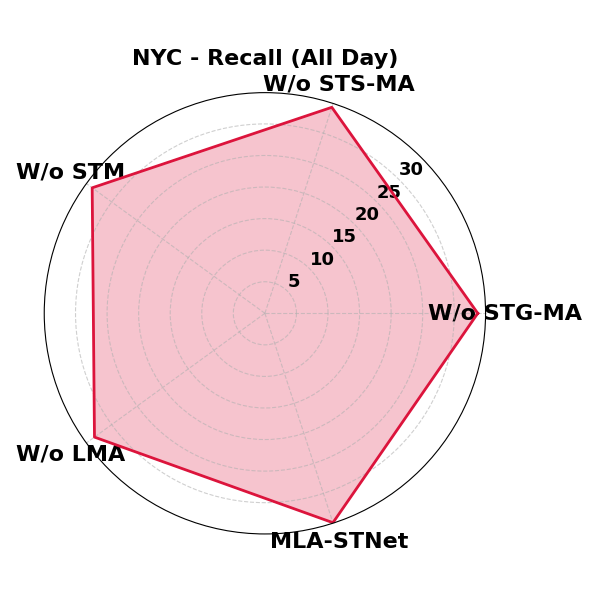}
        \caption{Recall comparison}
    \end{subfigure}
    \hfill
    \begin{subfigure}{0.32\textwidth}
        \includegraphics[width=\linewidth]{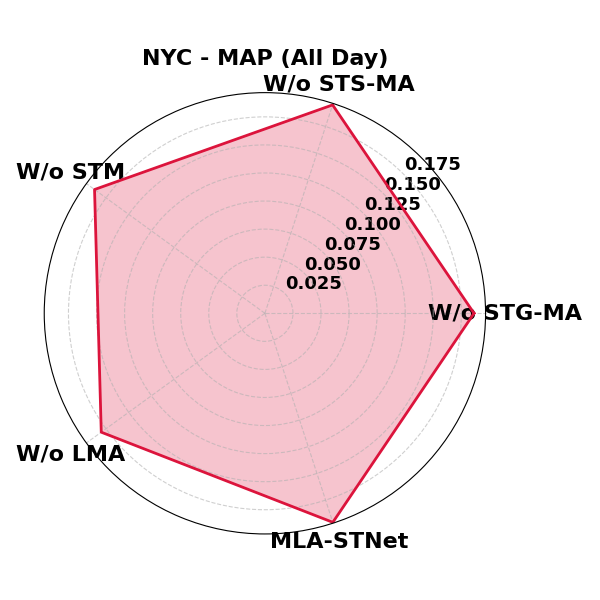}
        \caption{MAP comparison}
    \end{subfigure}
    \caption{Radar chart visualization of ablation results on the New York City (NYC) dataset under the All-day setting. Each subfigure compares the full \textsf{MLA-STNet} model with its four ablated variants (W/o STG-MA, W/o STS-MA, W/o STM, and W/o LMA) in terms of RMSE, Recall, and MAP.}
    \label{fig:ablation-New York City (NYC)}
\end{figure}

\begin{figure}[htbp!]
    \centering
    \begin{subfigure}{0.32\textwidth}
        \includegraphics[width=\linewidth]{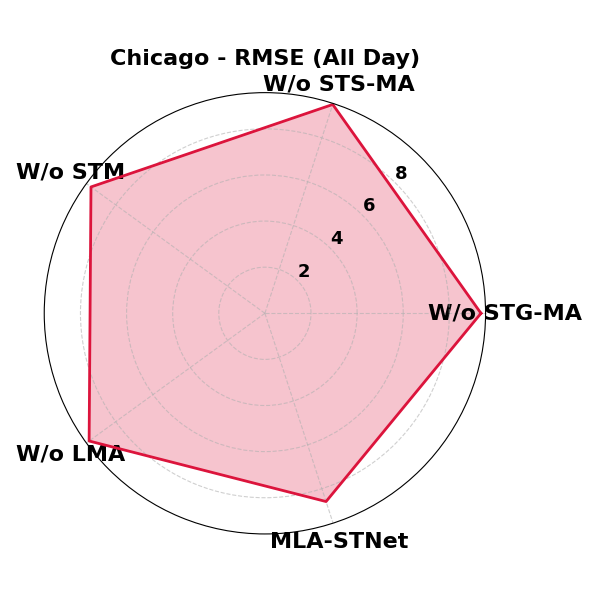}
        \caption{RMSE comparison}
    \end{subfigure}
    \hfill
    \begin{subfigure}{0.32\textwidth}
        \includegraphics[width=\linewidth]{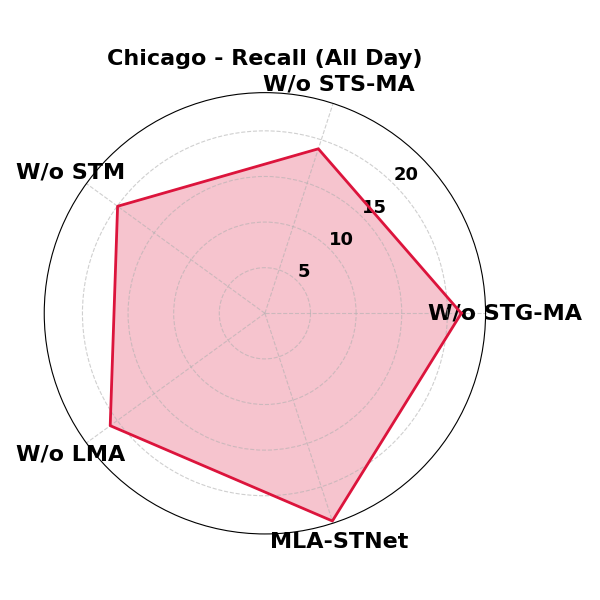}
        \caption{Recall comparison}
    \end{subfigure}
    \hfill
    \begin{subfigure}{0.32\textwidth}
        \includegraphics[width=\linewidth]{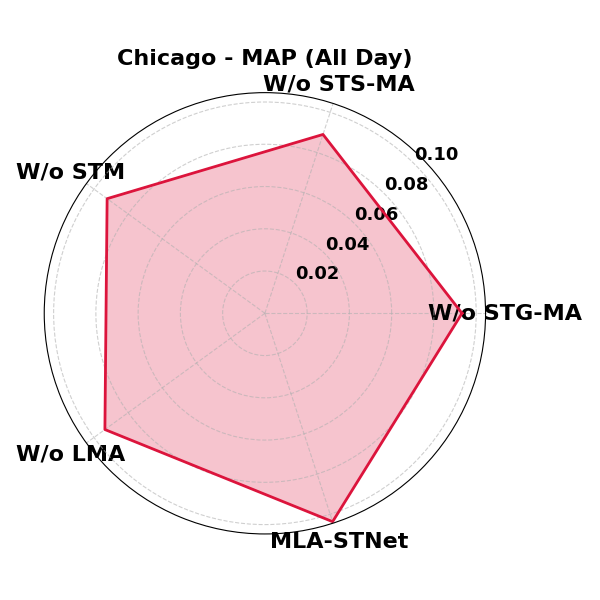}
        \caption{MAP comparison}
    \end{subfigure}
    \caption{Radar chart visualization of ablation results on the Chicago dataset under the \textbf{All-day} setting. The plots illustrate the performance differences between \textsf{MLA-STNet} and its four ablated versions in three evaluation metrics (RMSE, Recall, and MAP).}
    \label{fig:ablation-chicago}
\end{figure}

\paragraph{Quantitative Analysis of Ablation}  
The ablation results in Table~\ref{tab:ablation}, Fig.~\ref{fig:ablation-New York City (NYC)} and Fig.~\ref{fig:ablation-chicago} quantitatively and visually demonstrate the effectiveness of each proposed component in \textsf{MLA-STNet}. On the New York City (NYC) dataset, removing STG-MA causes RMSE to rise from 6.88 to 7.00 and MAP to drop from 0.1958 to 0.1863, showing that STG-MA effectively suppresses spatial noise and stabilizes temporal variations. Without STG-MA, attention becomes more dispersed across low-risk regions, increasing overall prediction error and reducing hotspot accuracy.  

Removing STS-MA mainly impacts recall performance. In the Chicago dataset, recall decreases from 23.94\% to 18.96\%, while RMSE rises from 8.59 to 9.52. This indicates that STS-MA enhances the model’s ability to capture heterogeneous semantic cues, such as road topology and accident distribution patterns, across cities. Without it, the model under-detects high-risk transition areas, leading to missed hotspots even when MAP remains similar (0.1037 → 0.0890).  

The STM component contributes to temporal stability. When STM is removed, RMSE increases from 6.88 to 7.10 in New York City (NYC) and from 8.59 to 9.32 in Chicago, suggesting that the model loses the ability to maintain smooth long-term temporal dependencies. Meanwhile, the LMA component strengthens local spatial discrimination. Its removal leads to a MAP decline from 0.1958 to 0.1802 in New York City (NYC) and from 0.1037 to 0.0936 in Chicago, indicating less precise hotspot localization.  

Overall, these quantitative results confirm that STG-MA reduces spatial dispersion, STS-MA improves cross-city semantic adaptability, STM stabilizes long-range temporal dependencies, and LMA enhances local hotspot precision. Their integration allows \textsf{MLA-STNet} to achieve the lowest RMSE (6.88 in New York City (NYC) and 8.59 in Chicago) and the highest recall (34.88\% and 23.94\%, respectively), highlighting the complementary nature of these components in improving both predictive accuracy and interpretability.

\begin{table}[htbp!]
\centering
\footnotesize
\caption{The contribution of each component for the \textsf{MLA-STNet}}
\label{tab:ablation}
\renewcommand{\arraystretch}{1.2}

\begin{tabular}{|c|l|c|c|c|c|c|c|}
\hline
\multirow{2}{*}{Dataset} & \multirow{2}{*}{Model} 
& \multicolumn{3}{c|}{All day} & \multicolumn{3}{c|}{High-frequency accident periods} \\
\cline{3-8}
& & RMSE & Recall & MAP & RMSE & Recall & MAP \\
\hline
\multirow{6}{*}{New York City (NYC)}
 & W/o STG-MA                    & 7.0019 & 33.72\% & 0.1863 & 6.2873 & 34.01\% & 0.1796 \\
 & W/o STS-MA                 & 7.0548 & 34.31\% & \textbf{0.1952} & 6.3236 & 34.11\% & 0.1760 \\
 & W/o STM             & 7.1023 & 33.82\% & 0.1876 & 6.5021 & 34.32\% & 0.1799 \\
 & W/o LMA            & 7.1348 & 33.37\% & 0.1802 & 6.4982 & 34.46\% & 0.1827 \\
 \rowcolor{gray!20}&  \textbf{\textsf{MLA-STNet}}                     & \textbf{6.8824} & \textbf{34.88\%} & 0.1958 & \textbf{6.4191} & \textbf{35.22\%} & 0.1834 \\

\hline
\multirow{6}{*}{Chicago}
 & W/o STG-MA                     & 9.3724 & 21.53\% & \textbf{0.0933} & 6.6925 & 23.32\% & 0.1169 \\
 & W/o STS-MA                 & 9.5245 & 18.96\% & 0.0890 & 6.9466 & 18.93\% & 0.1096 \\
 & W/o STM                 & 9.3245 & 19.96\% & 0.0923 & 6.6466 & 20.96\% & 0.1066 \\
 & W/o LMA            & 9.4242 & 20.96\% & 0.0936 & 6.6743 & 21.95\% & 0.1076 \\
 \rowcolor{gray!20}& \textbf{\textsf{MLA-STNet}}                     & \textbf{8.5855} & \textbf{23.94\%} & \textbf{0.1037} & 6.4491 & \textbf{23.13\%} & \textbf{0.1163} \\

\hline
\end{tabular}

\end{table}

\subsection{Efficiency Analysis}

To comprehensively evaluate the balance between accuracy and efficiency, we design a trade-off metric that integrates prediction error and computational cost into a single score. Specifically, we consider the Root Mean Squared Error (RMSE) as the measure of accuracy, while the efficiency is reflected by the inference time of a single forward pass ($T_{forward}$) and the time per batch ($T_{batch}$), and the experiments are conducted on the Chicago dataset. Formally, the trade-off score is defined as a weighted combination:
\begin{table}[ht]
\centering
\footnotesize
\caption{Trade-off analysis with RMSE in raw values, while $T_{forward}$ and $T_{batch}$ are normalized to [1, 10]. Lower score indicates better balance.}
\resizebox{\textwidth}{!}{%
\begin{tabular}{lcccc|c}
\hline
\textbf{Model} & \textbf{RMSE} & \textbf{$T_{forward}$ (norm.)} & \textbf{$T_{batch}$ (norm.)} & \textbf{Weights} & \textbf{Trade-off Score} \\
\hline
\rowcolor{gray!20} \textsf{MLA-STNet}    & 8.5855  & 1.137 & 1.401 & 50\%/25\%/25\% & \textbf{4.927} \\
MG-STNET     & 9.1195  & 10.000 & 10.000 & 50\%/25\%/25\% & 9.560 \\
GSNet        & 10.6302 & 1.759 & 2.915 & 50\%/25\%/25\% & 6.484 \\
TWCCNet      & 11.1277 & 1.811 & 3.573 & 50\%/25\%/25\% & 6.910 \\
ViT-traffic  & 9.4079  & 1.000 & 1.000 & 50\%/25\%/25\% & 5.204 \\
\hline
\end{tabular}%
}
\label{tab:tradeoff}
\end{table}
\begin{equation}
\text{Trade-off Score} = \alpha \times \text{RMSE} + \beta \times T_{\text{forward}} + \gamma \times T_{\text{batch}},
\end{equation}
where $\alpha, \beta, \gamma$ are non-negative weights that control the relative importance of accuracy and efficiency. 
In our experiments, we set $\alpha=0.5$, $\beta=0.25$, and $\gamma=0.25$ to emphasize the dominant role of accuracy while still penalizing high computational cost. 
This design ensures that the score captures both prediction quality and practical deployability of the models.

\begin{figure}[htbp!]
    \centering
    \includegraphics[width=0.75\textwidth]{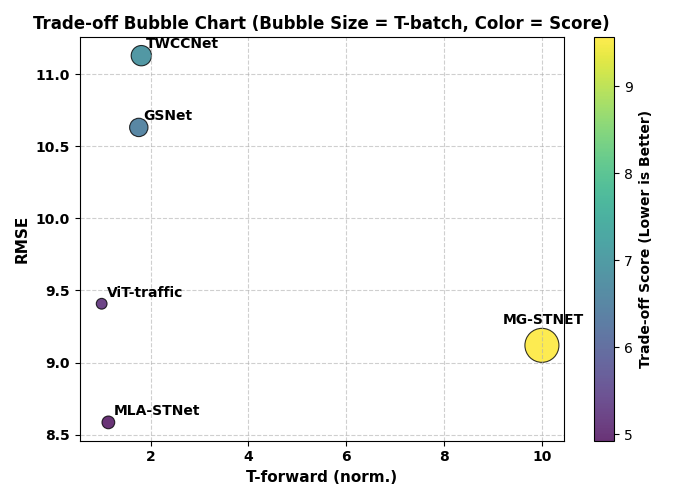}
    \caption{Trade-off bubble chart of different models on the New York City (NYC) dataset.}
    \label{fig:bubble_chart}
\end{figure}

Fig.~\ref{fig:bubble_chart} presents a trade-off bubble chart that jointly evaluates accuracy and efficiency across different models. The x-axis represents the forward inference time ($T_{\text{forward}}$), the y-axis indicates the prediction error (RMSE), the bubble size corresponds to the training time per batch ($T_{\text{batch}}$), and the color encodes the composite trade-off score, where a lower value implies better performance. From the figure, \textsf{MLA-STNet} achieves the lowest RMSE with relatively small computational cost, leading to the most favorable trade-off score. ViT-traffic benefits from very fast inference but suffers from higher error, while MG-STNet incurs heavy computational overhead despite moderate accuracy. TWCCNet and GSNet remain less competitive due to larger errors. Overall, the visualization highlights that \textsf{MLA-STNet} achieves the most balanced performance in terms of both accuracy and efficiency.

\subsection{Robustness Analysis}

\begin{table}[ht]
\centering

\caption{All-day robustness comparison under different noise levels on the Chicago dataset. Lower RMSE and higher Recall/MAP indicate better performance.}
\label{tab:robustness_all}
\renewcommand{\arraystretch}{1.15}

\begin{tabular}{c|>{\columncolor{gray!20}}c>{\columncolor{gray!20}}c>{\columncolor{gray!20}}c|ccc|ccc}
\hline
\multirow{2}{*}{\shortstack{Noise Level\\(multi-task)}} &
\multicolumn{3}{c|}{\cellcolor{gray!20}Ours (\textsf{MLA-STNet})} &
\multicolumn{3}{c|}{GSNet} &
\multicolumn{3}{c}{ViT-Traffic} \\
\cline{2-10}
 & RMSE & Recall & MAP & RMSE & Recall & MAP & RMSE & Recall & MAP \\
\hline
0.0 & 8.57 & 23.8\,\% & 0.103 & 12.65 & 18.8\,\% & 0.076 & 11.42 & 17.2\,\% & 0.075 \\
0.1 & 8.57 & 23.8\,\% & 0.103 & 12.80 & 18.6\,\% & 0.074 & 11.59 & 17.0\,\% & 0.073 \\
0.2 & 8.67 & 23.6\,\% & 0.102 & 12.96 & 18.3\,\% & 0.072 & 11.78 & 16.8\,\% & 0.071 \\
0.3 & 8.63 & 23.7\,\% & 0.103 & 13.18 & 18.0\,\% & 0.070 & 12.01 & 16.5\,\% & 0.069 \\
0.5 & 8.53 & 23.7\,\% & 0.103 & 13.45 & 17.7\,\% & 0.067 & 12.30 & 16.2\,\% & 0.066 \\
\hline
\end{tabular}%

\end{table}

\begin{table}[ht]
\centering

\caption{High-frequency accident periods robustness comparison under different noise levels on the Chicago dataset. Lower RMSE and higher Recall/MAP indicate better performance.}
\label{tab:robustness_high}
\renewcommand{\arraystretch}{1.15}

\begin{tabular}{c|>{\columncolor{gray!20}}c>{\columncolor{gray!20}}c>{\columncolor{gray!20}}c|ccc|ccc}
\hline
\multirow{2}{*}{\shortstack{Noise Level\\(multi-task)}} &
\multicolumn{3}{c|}{\cellcolor{gray!20}Ours (\textsf{MLA-STNet})} &
\multicolumn{3}{c|}{GSNet} &
\multicolumn{3}{c}{ViT-Traffic} \\
\cline{2-10}
 & RMSE & Recall & MAP & RMSE & Recall & MAP & RMSE & Recall & MAP \\
\hline
0.0 & 6.44 & 23.1\,\% & 0.116 & 10.92 & 19.1\,\% & 0.089 & 8.65 & 18.0\,\% & 0.087 \\
0.1 & 6.45 & 23.1\,\% & 0.115 & 11.10 & 18.9\,\% & 0.087 & 8.79 & 17.8\,\% & 0.085 \\
0.2 & 6.46 & 23.0\,\% & 0.114 & 11.31 & 18.7\,\% & 0.085 & 8.95 & 17.6\,\% & 0.083 \\
0.3 & 6.44 & 23.1\,\% & 0.114 & 11.55 & 18.4\,\% & 0.083 & 9.13 & 17.4\,\% & 0.081 \\
0.5 & 6.45 & 23.1\,\% & 0.116 & 11.82 & 18.1\,\% & 0.080 & 9.38 & 17.1\,\% & 0.078 \\
\hline
\end{tabular}%

\end{table}

\begin{figure}[htbp!]
    \centering
    \includegraphics[width=\textwidth]{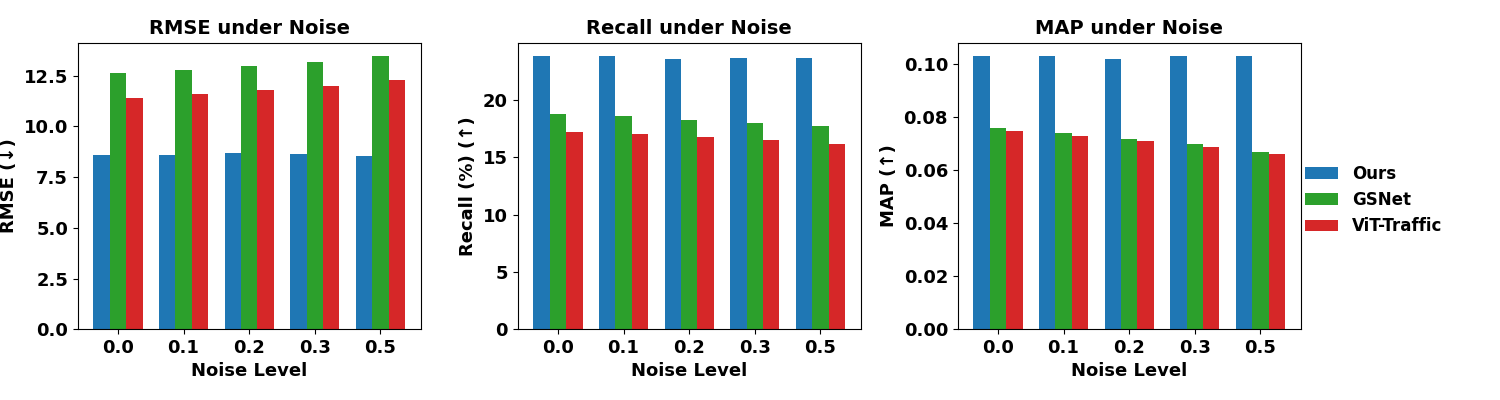} 
    \caption{All-day RMSE, Recall, and MAP under different noise levels on the Chicago dataset for \textsf{MLA-STNet} (ours), GSNet, and ViT-Traffic.}
    \label{noise}
\end{figure}
\begin{figure}[htbp!]
    \centering
    \includegraphics[width=\textwidth]{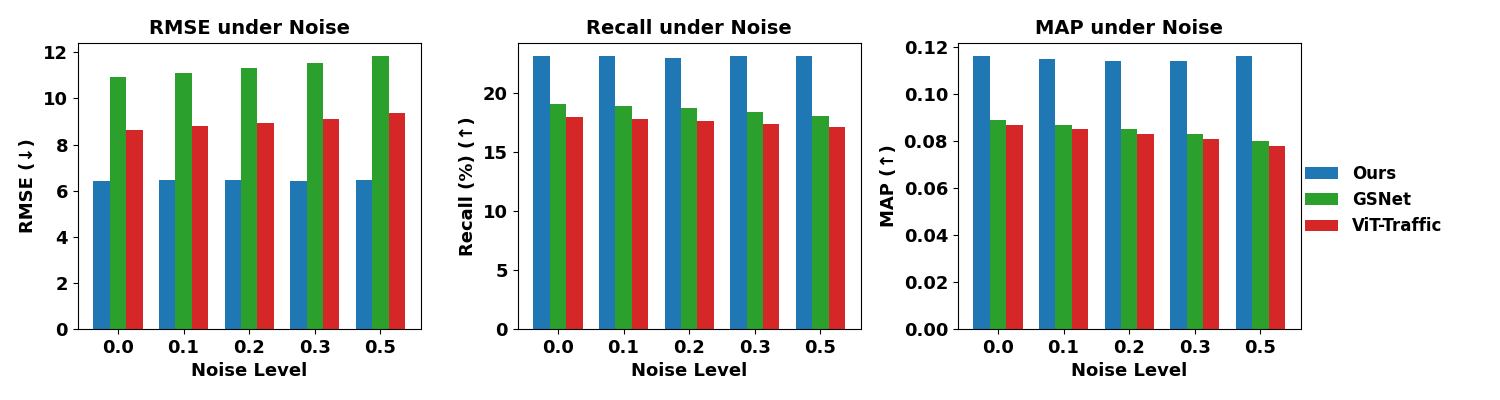} 
    \caption{High-frequency
accident periods RMSE, Recall, and MAP under different noise levels on the Chicago dataset for \textsf{MLA-STNet} (ours), GSNet, and ViT-Traffic.}
    \label{noiseHigh}
\end{figure}

Accident datasets are inherently noisy due to inconsistent reporting, missing entries, and random environmental variations. To assess the robustness of \textsf{MLA-STNet} under such uncertainty, we conducted experiments under the multi-task setting by adding Gaussian noise of varying intensities (ranging from 0.0 to 0.5) to the input features. The results for the All-day setting are reported in Table~\ref{tab:robustness_all}. Although all models exhibit a slight degradation as the noise level increases, \textsf{MLA-STNet} remains notably stable, with RMSE fluctuating only between 8.53 and 8.67, Recall consistently around 23.7\%, and MAP maintaining values between 0.102 and 0.103. In contrast, GSNet shows a steady decline in performance, with RMSE increasing from 12.65 to 13.45 and MAP decreasing from 0.076 to 0.067. Similarly, ViT-Traffic demonstrates higher sensitivity to noise, with its RMSE rising from 11.42 to 12.30 and MAP dropping from 0.075 to 0.066. These results confirm that \textsf{MLA-STNet} effectively preserves prediction accuracy and ranking consistency even when the input data are perturbed, indicating strong resistance to random noise under the full-day condition.

A similar trend is observed under the High-frequency accident periods, as shown in Table~\ref{tab:robustness_high}. \textsf{MLA-STNet} again demonstrates remarkable stability, with RMSE varying minimally between 6.44 and 6.46, Recall remaining steady at approximately 23.1\%, and MAP fluctuating only within 0.002. By contrast, GSNet’s RMSE rises from 10.92 to 11.82 and MAP declines from 0.089 to 0.080, while ViT-Traffic’s MAP drops from 0.087 to 0.078. These results highlight that \textsf{MLA-STNet} maintains robust and reliable forecasting capability even in high-risk temporal segments characterized by dense accident occurrences and higher data uncertainty. The robustness can be attributed to the model’s multi-task shared-parameter design, which allows consistent parameter learning across cities while enabling each task to adaptively learn its own feature representations. This structure enhances generalization and mitigates overfitting to local noise, ensuring stability in real-world cross-city accident prediction scenarios.
\section{Discussion and Implementation}\label{s6}
\paragraph{Discussion}

This study demonstrates the efficacy of \textsf{MLA-STNet} in addressing cross-city accident risk forecasting under heterogeneous, sparse, cyclical, and noisy conditions. Across two temporal settings, comprehensive all-day forecasting and high-frequency accident intervals, and two real-world datasets (New York City and Chicago), the model consistently surpasses competitive baselines. Compared with the strongest alternative, the multi-task variant achieves up to a 6\% reduction in RMSE, an 8\% improvement in Recall, and a 5\% gain in MAP. These outcomes substantiate the proposed unified cross-city framework, which jointly captures transferable temporal dynamics while maintaining city-specific semantics, thereby eliminating the need for isolated city-wise models.

A detailed examination of the results clarifies the mechanisms underlying these improvements. The Spatio–Temporal Geographical Mamba-Attention (STG-MA) module effectively suppresses unstable local fluctuations and enhances long-range temporal coherence, an essential property for modeling spatially clustered yet temporally sparse accident occurrences with recurrent commuting patterns. Concurrently, the Spatio–Temporal Semantic Mamba-Attention (STS-MA) module mitigates cross-city heterogeneity through shared parameters while preserving distinct semantic representations for each city (e.g., road topology, mobility, and meteorological context). This complementary design enables collaborative learning across urban domains while retaining regional specificity.

Despite these advantages, several challenges remain. Accident risks are spatially concentrated but temporally irregular, resulting in uneven label distributions. Although STG-MA alleviates overfitting to transient spikes, extreme sparsity can still cause local under- or over-smoothing in weak-signal regions. Moreover, variations in peak timing and amplitude across cities introduce phase misalignments that complicate joint temporal learning, even after relative-time normalization. Semantic imbalance further limits performance: New York City offers denser POI data and longer observation periods, whereas Chicago exhibits lower semantic richness and shorter records. While STS-MA alleviates such disparities, modality-induced degradation cannot be entirely eliminated. Additionally, constructing multi-view similarities and refining adaptive relations incur computational costs; although Mamba-based operators are efficient, real-time implementation requires careful management of graph scale, update frequency, and temporal window size.

From a broader systems perspective, the multi-task configuration offers a viable route toward collaborative accident prevention, allowing agencies to share transferable parameters while maintaining local data sovereignty. Scalable real-time decision support, however, depends on engineering refinements such as model pruning and knowledge distillation for compactness, risk-aware graph sparsification to prioritize high-impact regions, and distributed inference combining low-latency edge computation with periodic cloud synchronization. Given that accident alerts directly inform public interventions, interpretability and calibrated uncertainty estimation remain critical for ensuring transparency and accountability in policy decisions.

In conclusion, \textsf{MLA-STNet} marks a substantial step toward an integrated Cross-City Accident Prevention System. By coupling transferable temporal kernels with city-specific semantic spaces, the model achieves superior accuracy and robustness across diverse metropolitan settings. Future efforts should focus on enhancing semantic completeness, improving temporal synchronization, optimizing computational efficiency, and strengthening interpretability to support trustworthy and scalable deployment for urban safety governance.

\paragraph{Implementation}
\begin{figure}[htbp!]
    \centering
    \includegraphics[width=\textwidth]{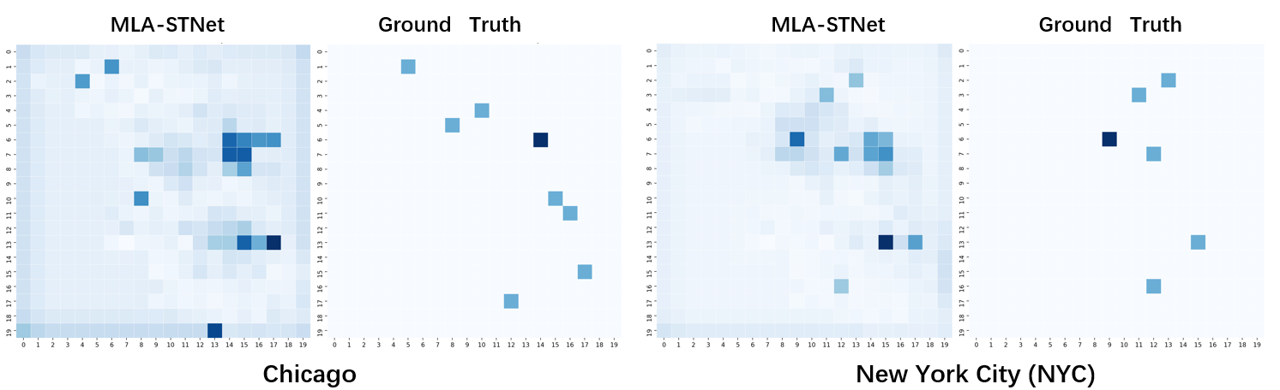}
    \caption{Comparison between \textsf{MLA-STNet} predictions and ground truth in Chicago and New York City (NYC) at the same time.}

    \label{sametime}
\end{figure}

Fig.~\ref{sametime} illustrates \emph{simultaneous} multi-city forecasting: at the same time slice, \textsf{MLA-STNet} produces a predicted risk field for Chicago and New York City in parallel (left subpanels for each city), alongside the corresponding sparse ground-truth realizations (right subpanels). The juxtaposition highlights what the multi-task design is intended to do in one pass: share temporal kernels across cities while respecting city-specific semantics, yielding smooth but compact risk patches that reflect the distributional footprint of each city’s observed incidents without fabricating extraneous structure. Rather than fitting each jurisdiction separately and then reconciling outputs post hoc, the model infers both cities jointly under a shared parameterization with city-local semantic spaces, so the two predicted fields are calibrated on a common scale and directly comparable \emph{at the same moment}.

In operations, this single-shot, cross-city inference is consequential. Transport agencies can coordinate deployments when both predicted fields indicate concurrent persistence, while avoiding overreaction where the mask-driven aggregation shows weak or uncertain signals. Urban planners read stability across adjacent cells to prioritize corridor-level treatments within each city, yet the shared calibration allows evidence from one jurisdiction to transfer credibly to its peer when semantic structure is analogous. EMS and first responders exploit coherent predicted patches for pre-positioning across cities during the same hour, improving joint readiness for regional shocks. Policy teams, insurers, and public dashboards benefit from aligned indicators that tie both cities’ forecasts to common KPIs without city-specific recalibration. In short, the figure demonstrates that \textsf{MLA-STNet} converts extremely sparse, heterogeneous observations into synchronized, comparable risk surfaces for multiple cities in a single forward pass, enabling coordinated planning, targeted enforcement, and real-time resource allocation across jurisdictions.

\section{Conclusion}\label{s7}

This study presents \textsf{MLA-STNet}, a unified framework that forms the analytical foundation of a \textit{Cross-City Accident Prevention System} within the broader context of Intelligent Transportation Systems (ITS). By coupling the Spatio–Temporal Geographical Mamba-Attention (STG-MA) and Spatio–Temporal Semantic Mamba-Attention (STS-MA) modules, the proposed model effectively captures clustered, sparse, cyclical, and noisy accident patterns, while mitigating cross-city heterogeneity caused by diverse infrastructures, reporting standards, and data governance practices. Through a shared-parameter multi-task architecture, \textsf{MLA-STNet} enables collaborative learning of transferable spatio–temporal dynamics across cities, while preserving local semantic structures, thereby supporting coordinated and interpretable accident forecasting across heterogeneous metropolitan systems.

Empirical evaluations on large-scale datasets from Chicago and New York City demonstrate that \textsf{MLA-STNet} achieves substantial improvements over state-of-the-art baselines, including up to a 6\% reduction in RMSE, 8\% increase in Recall, and 5\% gain in MAP. More importantly, these results confirm that knowledge transfer across cities enhances predictive reliability and stability under noisy and incomplete data conditions. Such capability is crucial for real-world ITS applications, where multi-jurisdictional coordination and data-driven early-warning systems are essential for proactive traffic management and resource allocation. The proposed framework thus provides not only a methodological advancement in spatio–temporal modeling but also a scalable operational pathway toward the realization of an intelligent \textit{Cross-City Accident Prevention System}. Looking forward, future research will focus on three major directions: (i) developing adaptive and causal graph structures to dynamically capture evolving mobility, social, and policy interactions; (ii) extending the framework to federated and privacy-preserving learning environments that enable secure collaboration among jurisdictions; and (iii) integrating real-time inference pipelines to support early-warning and decision-making in large-scale ITS operations. Ultimately, \textsf{MLA-STNet} bridges the gap between theoretical spatio–temporal modeling and practical ITS deployment, offering a cooperative, interpretable, and resilient foundation for cross-city accident prevention, advancing the vision of safer, smarter, and more sustainable urban mobility.







\bibliographystyle{elsarticle-harv}
\bibliography{cas-refs}   

\end{document}